\documentclass{amsart}
\usepackage{graphicx}

\usepackage[utf8]{inputenc} 
\usepackage[T1]{fontenc}    
\usepackage{hyperref}       
\usepackage{url}            
\usepackage{booktabs}       
\usepackage{amsfonts}       
\usepackage{nicefrac}       
\usepackage{microtype}      
\usepackage{xcolor}         

\usepackage[margin = 1in]{geometry}

\usepackage{amsmath}
\usepackage{amssymb}
\usepackage[foot]{amsaddr}
\usepackage{mathtools}
\usepackage{amsthm}
\usepackage{multirow}
\usepackage{soul}
\usepackage[capitalize,noabbrev]{cleveref}
\usepackage{natbib}
\usepackage{dsfont}
\usepackage{float}
\usepackage{combelow}
\usepackage{caption}
\usepackage{subcaption}
\usepackage{stfloats}
\usepackage{listings}
\usepackage{comment}
\usepackage[capitalize,noabbrev]{cleveref}
\usepackage{todonotes}

\DeclareMathOperator*{\minimize}{minimize}
\DeclareMathOperator*{\argmin}{argmin}
\newcommand*\xbar[1]{%
  \hbox{%
    \vbox{%
      \hrule height 0.5pt 
      \kern0.5ex
      \hbox{%
        \kern-0.1em
        \ensuremath{#1}%
        \kern-0.1em
      }%
    }%
  }%
}

\def\INCLUDEFIGS{}

\title{Comparing EPGP Surrogates and Finite Elements Under Degree-of-Freedom Parity}

\author{Obed~Amo}
\email[OA]{oamo@odu.edu}

\author{Samit~Ghosh}
\email[SG]{s2ghosh@odu.edu}

\address[OA, SG, MP]{Department of Mathematics and Statistics, Old Dominion University, Norfolk, VA 23529, USA}

\author{Markus~Lange-Hegermann}
\email[MLH]{markus.lange-hegermann@th-owl.de}
\address[MLH]{Department of Electrical Engineering and Computer Science, OWL University of Applied Sciences and Arts, 32657 Lemgo, Germany}

\author{Bogdan~Rai\cb{t}\u{a}}
\email[BR]{bogdan.raita@georgetown.edu}
\address[BR]{Department of Mathematics and Statistics, Georgetown University, Washington, DC 20057, USA}

\author{Michael~Pokojovy}
\email[MP]{mpokojovy@odu.edu {\rmfamily (Corresponding author)}}

\begin{document}

\begin{abstract}
    We present a new benchmarking study comparing a boundary-constrained Ehrenpreis--Palamodov Gaussian Process (B-EPGP) surrogate with a classical finite element method combined with Crank--Nicolson time stepping (CN-FEM) for solving the two-dimensional wave equation with homogeneous Dirichlet boundary conditions. The B-EPGP construction leverages exponential-polynomial bases derived from the characteristic variety to enforce the PDE and boundary conditions exactly and employs penalized least squares to estimate the coefficients. To ensure fairness across paradigms, we introduce a degrees-of-freedom (DoF) matching protocol. Under matched DoF, B-EPGP consistently attains lower space-time $L^2$-error and maximum-in-time $L^{2}$-error in space than CN-FEM, improving accuracy by roughly two orders of magnitude. \\[5pt]

\end{abstract}
\maketitle

\section{Introduction}\label{sec:intro}

\subsection*{Background}
    
Partial differential equations (PDEs) form the backbone of mathematical modeling in science and engineering. Notably, the majority of rational mechanics relies on PDEs to establish macroscropic models of continua such as elastic bodies and other structures, fluids, electromagnetic fields, etc. For decades, deterministic numerical methods have been the dominant tool for approximating PDE solutions with guaranteed accuracy. Nevertheless, their computational cost scales poorly with mesh resolution and simulation time. They are also only applicable to well-posed problems, necessitating the unique solvability and continuous dependence on the data in abstract functional spaces.

In recent years, probabilistic approaches based on machine learning (ML) have emerged as powerful alternatives to classical methods. To approximate PDE solutions directly from data, frameworks like physics-informed neural networks (PINNs)~\citep{raissi2019physics} use deep learning models that incorporate the governing PDEs into the regression loss function and learn solutions from data; 
Gaussian process (GP) regression~\citep{Pfortner2022PhysInfGP} is a Bayesian non-parametric approach that models distributions over functions to perform regression with uncertainty quantification;  operator-based learning approaches~\citep{Kovachki2023NeuralOperator,li2024physics} are also popular nowadays. GP regression, in particular, has been extended to incorporate linear operator constraints that fit the observed data, making it possible to encode PDE structure directly into the prior~\citep{jidling2017linearly,LangeHegermann2018}. The Ehrenpreis--Palamodov fundamental principle establishes that solutions of constant-coefficient linear PDEs can be represented as superpositions of exponential functions, providing a theoretical foundation for operator-learning approaches~\citep{ehrenpreis2011fourier,palamodov1970linear}. H\"ark\"onen et al.~\citep{harkonen2023gaussian} introduced Ehrenpreis--Palamodov Gaussian process (EPGP) priors, which leverage the aforementioned fundamental principle to construct exact solution priors for linear PDEs with constant coefficients, enabling inference from noisy or partial data. Also these methods are computationally cost effective, while achieving higher accuracy than classical numerical methods. 

\subsection*{Contribution}
  
These methods from machine learning facilitate rapid evaluation and follow Bayesian principles to allow calibrated uncertainty quantification. Thereby, they often replace rigorous {\it a priori} error analysis and stability guarantees that are typical for traditional solvers with probabilistic arguments that focus on average cases and average errors. Furthermore, fair comparisons oftentimes remain elusive because the two paradigms rely on fundamentally different assumptions about accuracy, complexity, and computational cost. Thus, the aim of the present paper is to compare the two paradigms quantitatively, in a fair manner. We benchmark both approaches on the classic wave equation in two dimensions subject to homogeneous Dirichlet boundary conditions.
  
As an archetypal numerical method, we employ the traditional piecewise linear $H^{1}$-finite element discretization of the spatial domain combined with Crank-Nicolson time integrator, a well-established implicit scheme that balances accuracy and stability~\citep{brenner2008mathematical,Johnson2009}. On the machine learning side, we use the EPGP method, which enforces the PDE structure and the boundary conditions to be satisfied \emph{exactly} directly in the GP prior. 
In this formulation, realizations satisfy the wave equation and boundary conditions in distribution~\citep{besginow2022constraining,harkonen2023gaussian,jidling2017linearly,LangeHegermann2018,Li2025}, yielding surrogate solutions with quantified uncertainty. For the wave equation~\eqref{eq:2D_WAVE_EQUATION}, the GP prior is chosen for ensuring compatibility with the PDE constraints~\citep{besginow2022constraining,harkonen2023gaussian,Li2025}. The resulting posterior mean provides a surrogate solution, while the posterior covariance quantifies uncertainty---a feature unavailable in classical FEM. Extensions of this framework have tackled boundary value problems~\citep{Gulian2022}, optimal control problems~\citep{besginow2025linear}, and high-frequency or multiscale PDEs~\citep{pmlr-v235-dalton24a,fang2023solving,Henderson2023}.  

To ensure fairness, we introduce a benchmarking framework based on degrees of freedom (DoF) matching. For FEM, the DoF scales with the number of spatial nodes and time steps, while for EPGP it is defined via the posterior trace of the regression operator, known as influence or hat matrix in statistics. This matching allows us to compare both methods on equal footing, evaluating accuracy, stability, computational efficiency, and error-complexity tradeoffs.  

Our results demonstrate that EPGP achieves higher accuracy than FEM by two orders of magnitude when matched for DoF, while providing near-instantaneous evaluations/predictions once trained. Moreover, unlike classical schemes, EPGP naturally quantifies uncertainty in data-driven fashion, offering insights into model reliability. Additionally, as a meshless approach, EPGP is not prone to the curse of dimensionality unlike its prominent grid-based competitors. Taken together, these findings highlight the complementary strengths of deterministic discretizations and probabilistic surrogates. While FEM remains the workhorse for robust and general-purpose PDE simulation, EPGP and related methods offer new opportunities for higher accuracy, faster evaluation, and uncertainty quantification.

This work presents a systematic benchmarking study between deterministic numerical solvers and probabilistic learning-based approaches for partial differential equations (PDEs). Building on~\cite{huang2024gaussian}, we propose a boundary-constrained Ehrenpreis--Palamodov Gaussian Process (B-EPGP) formulation that enforces PDE structure exactly while allowing efficient inference from limited data. Unlike conventional numerical schemes, the proposed method offers exact PDE satisfaction, uncertainty quantification, and rapid evaluation once trained. Our contribution lies in: 
\begin{enumerate}
    \item constructing a deterministic B-EPGP basis that satisfies both the governing wave equation and Dirichlet boundary conditions;
    
    \item developing a regularized least-squares fitting framework for stable coefficient estimation; 
    
    \item introducing a degree-of-freedom (DoF) matching criterion that enables fair comparison with a Finite Element Crank–Nicolson (CN-FEM) solver.
\end{enumerate}
The results demonstrate that B-EPGP achieves accuracy that is two orders of magnitude higher than that of CN-FEM under matched DoF, with significantly reduced computation time and built-in uncertainty quantification.

\subsection*{Relevance} 

The relevance of this work stems from the growing need for efficient and interpretable solvers for PDE-governed systems in science, engineering, quantitative finance, etc. Traditional numerical solvers, such as finite element and finite difference methods, offer stability and convergence guarantees but often suffer from high computational costs from fine mesh resolutions and long-time simulations. Meanwhile, emerging machine learning models, including physics-informed neural networks (PINNs) and operator learning frameworks, show promise in approximating PDE solutions directly from data but typically lack theoretical guarantees and rigorous error analysis. The Ehrenpreis--Palamodov Gaussian Process framework bridges these paradigms by embedding the analytical structure of PDEs within a probabilistic model. By comparing B-EPGP against CN-FEM under controlled conditions, our study provides insights into the trade-offs between deterministic accuracy, computational efficiency, and probabilistic uncertainty quantification---a highly relevant aspect of modern scientific computing.
    
\subsection*{Problem setup}

Our work addresses this gap through a careful benchmarking study that contrasts a classical variational solver with a probabilistic surrogate grounded in PDE structure. As a canonical test case, we consider the two-dimensional linear wave equation with constant coefficients on a rectangular domain \(\Omega = (0, L_{1}) \times (0, L_{2}) \subset \mathbb{R}^{2}\) subject to homogeneous Dirichlet boundary conditions:
\begin{equation}
    \label{eq:2D_WAVE_EQUATION}
    \begin{cases}
        u_{tt}(x, y,t) = c^{2}\left(u_{xx}(x, y, t)+ u_{yy}(x, y, t)\right), & (x, y) \in \Omega, \,t > 0, \\
        u(x, y, t) = 0, & (x, y) \in \partial \Omega,\, t>0,\\
        u(x, y, 0) = u_{0}(x, y), \quad u_{t}(x, y, 0) = v_{0}(x, y), &(x,y)\in\Omega,
    \end{cases}
\end{equation}
where \(c > 0\) is the wave speed, \(u_{0}\) is the prescribed initial displacement, and \(v_{0}\) is the initial velocity field.

The goal is to construct accurate surrogate and numerical approximations of $u(x,y,t)$ satisfying the above PDE and boundary conditions. The following two methods are considered: 
\begin{enumerate}
    \item[(i)] the boundary-constrained Ehrenpreis--Palamodov Gaussian Process (B-EPGP)~\citep{huang2024gaussian}, which uses exact analytical basis functions derived from the PDE’s characteristic variety; 
    
    \item[(ii)] the Finite Element Crank–Nicolson (CN-FEM) method, a well-established variational approach based on the method of lines ensuring second-order accuracy in both space and time.
\end{enumerate}
The wave equation is a fundamental model for wave propagation and an excellent benchmark for comparing numerical and learning-based solvers.
In fact, the wave equation is a canonical benchmark for numerical methods because its dynamics are well understood, the solution is smooth for smooth data and accurate discretizations are known. Our goal is to use this equation as a controlled setting to compare classical numerical schemes with machine learning methods based on Gaussian processes, with the goal of assessing accuracy, stability, and efficiency under matched degrees of freedom.
 
\subsection*{Previous work} 

Recent years have witnessed substantial interest in integrating machine learning with PDE solvers. Physics-Informed Neural Networks (PINNs)~\cite{raissi2019physics}, recent contributions such as Gaussian Process regression with PDE constraints~\citep{harkonen2023gaussian,jidling2017linearly}, and neural operator approaches~\citep{Kovachki2023NeuralOperator} have significantly advanced the integration of machine learning techniques with the modeling of physical systems. While PINNs incorporate PDE residuals into the neural loss, their optimization procedure is computationally intensive and prone to convergence issues. Gaussian Process (GP) approaches offer uncertainty quantification and analytical structure but often require linearity assumptions. The Ehrenpreis--Palamodov theorem~\citep{ehrenpreis2011fourier,palamodov1970linear} provides a foundation for constructing exact solution priors for linear PDEs with constant coefficients, enabling the development of EPGP priors~\citep{harkonen2023gaussian}. Extensions to boundary value problems~\citep{huang2024gaussian} and controlled systems~\citep{besginow2025linear} have recently been proposed. However, a rigorous comparison between such probabilistic PDE solvers and classical numerical schemes remains largely unexplored. Our work addresses this gap by benchmarking B-EPGP against CN-FEM using a systematic DoF-matched framework.

\subsection*{Outline}
    
The remainder of this paper is organized as follows. Section~\ref{sec:method} presents the methodological details of both the proposed B-EPGP approach and the finite element method paired with Crank–Nicolson integration we implemented in this paper. Section~\ref{sec:experiments} describes the numerical experiment design, including benchmark configurations and evaluation metrics. Section~\ref{sec:results} discusses the quantitative and qualitative aspects of the empirical results, followed by Section~\ref{sec:conclusion}, which summarizes the conclusions of our work and outlines potential future research directions.

\section{Methodology}\label{sec:method}

In this study, we compare two distinct approaches for solving the two-dimensional wave equation: the boundary Ehrenpreis–Palamodov Gaussian process (B-EPGP) technique~\citep{huang2024gaussian} and the classical Finite Element Method (FEM) with Crank–Nicolson time stepping~\citep{ern2004theory, larson2013finite}. Both frameworks are designed to ensure fairness through degrees-of-freedom (DoF) matching and are implemented under identical boundary and initial conditions to evaluate performance in terms of accuracy, stability, and computational cost.

The B-EPGP method~\citep{huang2024gaussian} extends the Ehrenpreis--Palamodov theorem~\citep{palamodov1970linear} and the EPGP algorithm~\citep{harkonen2023gaussian} to construct exact solution bases for linear partial differential equations with constant coefficients that also satisfy boundary constraints. For the wave equation \eqref{eq:2D_WAVE_EQUATION}, the characteristic variety determines the set of admissible exponential-polynomial functions, which are modified to enforce homogeneous Dirichlet conditions. The resulting B-EPGP basis functions thus satisfy the governing PDE and boundary conditions \textit{exactly} or \textit{symbolically}, leading to high accuracy and stability without requiring numerical discretization of the differential operator. The coefficients of the basis expansion are obtained through a regularized least-squares regression with the usual ridge penalty~\citep{hoerl1970ridge}, ensuring robustness against sampling noise and numerical instability. This framework maintains the interpretability of Gaussian processes while achieving the computational efficiency of a deterministic solver.

The classical Finite Element Method (FEM) combined with Crank–Nicolson time stepping serves as a deterministic benchmark. The spatial domain is discretized using linear triangular $H^{1}$-elements, leading to a semi-discrete system of equations with properly assembled mass and stiffness matrices. The Crank–Nicolson scheme provides second-order accuracy for the homogeneous wave equation in both space and time on sufficiently regular solutions, i.e.,
\begin{equation*}
    \|u - u_{h, \Delta t}\|_{L^{\infty}((0, T), L^{2}(\Omega))} = \mathcal{O}\big((\Delta t)^{2} + h^{2}\big) \text{ as } \Delta t, h \to 0,
\end{equation*}
and is unconditionally stable, i.e., no Courant--Friedrichs--Levy-type relation between the time step $\Delta t$ and the space step $h$ is required for convergence. This implicit formulation enables larger time steps without compromising stability, making FEM a robust and well-established numerical approach for time-dependent PDEs.

\subsection{B-EPGP}

Linear partial differential equations with constant coefficients posses a rich mathematical structure that can be exploited for constructing exact solution methods. The foundation of our approach rests on the Ehrenpreis--Palamodov theorem~\citep{harkonen2023gaussian, huang2024gaussian}, which characterizes the solution space of such equations in terms of exponential-polynomial functions. Consider a linear PDE with constant coefficients of the form $A (\boldsymbol\partial)u(\boldsymbol{x}) = 0$, where \(A \in \mathbb{C}[\boldsymbol{\partial}]\) and $\boldsymbol{\partial}=(\partial_{1}, \dots, \partial_{n})$, is the partial differential operator and \({\boldsymbol{x}}= (x_{1}, \dots, x_{n}) \in \mathbb{R}^{n} \). The characteristic variety of $A$ is defined by \(\mathcal{V} = \{\boldsymbol z\in \mathbb{C}^{n} : A(\boldsymbol z) = 0\}\), where \(A(\boldsymbol z)\) is obtained by substituting \(\partial_{j} \mapsto z_{j} \) into the differential operator. The Ehrenpreis--Palamodov theorem establishes the solutions to linear PDEs can be represented using exponential-polynomial functions whose frequencies lie in this characteristic variety. The original EPGP approach~\citep{harkonen2023gaussian} leverages the Ehrenpreis--Palamodov theorem to construct exact solutions for linear PDEs.

If the characteristic variety $\mathcal{V}$ has no multiplicities, solutions of \(A (\boldsymbol\partial)u(\boldsymbol x) = 0\) can be approximated with solutions of the form \(u(\boldsymbol x) = \sum_{j=1}^{r} w_{j}e^{\boldsymbol x\cdot \boldsymbol z_{j}}\), where \(\boldsymbol z_j \in\mathcal{V}\) are frequencies from the characteristic variety and \(w_{j}\) are coefficients to be determined from the data. Due to the linearity of the PDE, any linear combination of the exponential functions with frequencies from \(V\) automatically satisfy the differential equation exactly. The original EPGP approach~\citep{harkonen2023gaussian} treats the coefficients \(w_{j}\) as unknown probabilistic quantities giving rise to a Gaussian process formulation. To simplify our formulation and allow a more direct comparison to deterministic numerical methods, we ignore the probabilistic approach here and only use a direct regression method that computes the mean functions of the Gaussian process. 
While EPGP essentially produces solutions in full space, practical applications involve boundary conditions that constrain the solution on the domain boundary \(\partial\Omega\). The goal of~\citep{huang2024gaussian} is to construct basis functions that simultaneously satisfy both the differential equation and prescribed boundary constraints. For rectangular domains and Dirichlet/Neumann boundary conditions, this leads to a natural spectral decomposition, as we explain below.

Following~\citep{huang2024gaussian}, the B-EPGP method proceeds with an algebraic calculation aiming to modify exponential solutions with frequencies from the characteristic variety to satisfy boundary constraints. By explicit computation we can see that the wave equation \eqref{eq:2D_WAVE_EQUATION} with the differential operator \(A(\boldsymbol\partial) = {\partial^2_t} - c^{2} \left({\partial^2_x}+{\partial^2_y}\right) \)
has the characteristic variety \(\mathcal{V} = \{ (z_1,z_2,z_3) \in \mathbb{C}^{3} :z_3^{2} = c^{2} (z_1^2 + z_2^2) \}\).

Starting with the exponential solutions \( e^{z_3t + z_{1}x + z_{2}y} \) such that \( (\xi_{0}, \xi_{1}, \xi_{2})\in V \), the B-EPGP algorithm constructs linear combinations that satisfy the given boundary conditions. For rectangular domains with Dirichlet conditions, \citep{huang2024gaussian} calculates the B-EPGP basis 
\begin{equation*}
    \label{B_EPGP_BASIS_RECT}
    \exp\left({\pm c\sqrt{-1} \sqrt{\frac{j^{2}}{L_1^2} +\frac {k^{2}}{L_2^2}} \,\pi t} \right)\sin{\left( \frac{j\pi x}{L_{1}}  \right)}  \sin{\left( \frac{k\pi y}{L_{2}}  \right)}
\end{equation*}
for \(j, k \in \mathbb{Z}\). For the Neumann boundary condition, the sine functions are replaced by cosine functions. In this case, we retrieve classic Fourier series methods in two dimensions, which is due to the simple geometry of the rectangular domain. For different polygons, such as a generic triangle, B-EPGP produces continuously indexed bases with trainable frequencies from the characteristic variety $\mathcal V$.

Assuming zero initial velocity \(u_t(\cdot,0)={0}\) and expressing the complex exponential in the Cartesian form using Euler's identity \(\exp({\sqrt{-1}\omega t}) = \cos{(\omega t)} + \sqrt{-1} \sin{(\omega t)}\), the initial condition \(u_t(\cdot,0) = 0\) eliminates the sine term, leaving only the cosine component in the temporal basis. The basis functions are
\begin{equation*}
    \label{B_EPGP_BASIS_RECT_WAVE}
    \varphi_{j, k}(x, y,t) = \sin{\left( \frac{j\pi x}{L_{1}}  \right)}  \sin{\left( \frac{k\pi y}{L_{2}}  \right)} \cos{(\omega_{j,k} \,t)},
\end{equation*}
where \(\omega_{j,k} = c \sqrt{ (j\pi/L_{1})^2 +  (k\pi/L_{1})^2 }\). These \(\varphi_{j, k}(x, y,t)\)'s build a basis of the solution space of our PDE, including boundary conditions. We employ the ridge regression~\citep{hoerl1970ridge} approach to obtain a linear combination over a finite subset of these basis elements. This ensures that that our approximate solutions satisfy both the PDE and boundary conditions in~\eqref{eq:2D_WAVE_EQUATION} \textit{exactly}.

We seek to approximate a solution to \eqref{eq:2D_WAVE_EQUATION} with the finite-dimensional ansatz
\begin{equation*}
    \label{FINITE_DIMENSIONAL_APPROXIMATION}
    u_{N}(x, y, t) = \sum_{j=1}^{N}\sum_{k=1}^{N} w_{j,k} \,\varphi_{j,k}(x, y, t).
\end{equation*}
Since the problem \eqref{eq:2D_WAVE_EQUATION} is well-posed, the coefficients \( \{w_{j,k} \}\) can be determined by fitting to the initial condition $u(x,y,0)=u_0(x,y)$ for $(x,y)\in \Omega$:
\begin{equation*}
    \label{FINITE_DIMENSIONAL_APPROXIMATION_IC}
    u_{N}(x, y, 0) = \sum_{j=1}^{N}\sum_{k=1}^{N} w_{j,k} \,\varphi_{j,k}(x, y, 0).
\end{equation*}

Instead of fitting over the entire domain $\Omega$, we discretize the problem as a standard linear regression task by randomly selecting \(m\) spatial points \(S = \{ (x_{i}, y_{i}) \}_{i=1}^{m} \subset \Omega\). These points are generated using a Latin hypercube design (LHD)~\cite{damblin2013numerical,mckay1979comparison}, which provides an efficient and space-filling sampling strategy. Compared to using uniformly sampled points or an equispaced grid, the LHD achieves similar representational accuracy of the spatial domain with fewer samples, thereby reducing computational cost while maintaining good coverage and low correlation among sample locations~\cite{fang2006lhs}.
This novel addition to the EPGP methodology allows a more direct comparison to numerical approaches.

At each spatial grid point \((x_{i}, y_{i})\), we observe 
\begin{equation*}
    \label{MODEL_DATA}
    u_{i} = u_0(x_{i}, y_{i}) + \varepsilon_{i}, \quad i = 1, \dots,m
\end{equation*}
where \(u_{i}\) represents the observed initial displacement data at a point $(x_i,y_i)$ and \(\varepsilon_{i} \sim N(0, \sigma^{2})\) are independent random disturbances. Evaluating the basis function at the grid points yields the linear regression model
\begin{equation*}
    \label{REGRESSION_MODEL}
    \boldsymbol{u} = \boldsymbol{\Phi w} + \boldsymbol{\varepsilon}
\end{equation*}
where \( \boldsymbol{u} = [u_0(x_{1}, y_{1}), \dots, u_0(x_{m}, y_{m})]^{T} \in \mathbb{R}^{m}\) contains the observed initial condition values, the vector of coefficients
\(\boldsymbol{w} = [w_{1,1}, w_{1,2}, \dots, w_{N,N}]^{T} \in \mathbb{R}^{N^{2}}\) denotes the weights of the model,  $\boldsymbol{\Phi}\in\mathbb R^{m\times N^2}$ is the design matrix with entries
\begin{equation*}
    \Phi_{i, (j,k)} =  \sin{\left( \frac{j\pi x_{i}}{L_{1}}  \right)}  \sin{\left( \frac{k\pi y_{i}}{L_{2}}  \right)},
\end{equation*}
and \(\boldsymbol{\varepsilon} \in \mathbb{R}^{m}\) represents the error vector.

Under the Gaussian noise assumption $\boldsymbol{\varepsilon} \sim \mathcal{N}(0, \sigma^{2}\boldsymbol{I})$, the log-likelihood of $\boldsymbol{w}$ given the data is
\begin{equation*}
    \log p(\boldsymbol{u}\,|\,\boldsymbol{w}) 
    \propto -\tfrac{1}{2\sigma^{2}}\|\boldsymbol{\Phi w} - \boldsymbol{u}\|_{2}^{2}.
\end{equation*}
If, in addition, we impose a Gaussian prior $\boldsymbol{w} \sim \mathcal{N}(\mathbf{0}, \lambda \sigma^{2}\boldsymbol{I})$ while assuming independence from the error term, the negative log-posterior (up to an additive constant) becomes
\begin{equation*}
    \tfrac{1}{2\sigma^{2}}\|\boldsymbol{\Phi w} - \boldsymbol{u}\|_{2}^{2}
    + \tfrac{\lambda}{2\sigma^{2}}\|\boldsymbol{w}\|_{2}^{2}.
\end{equation*}
Maximizing the posterior, or equivalently minimizing its negative, yields the regularized least-squares problem
\begin{equation}
    \label{MINIMIZATION_PROBLEM}
    \minimize_{\boldsymbol{w} \in \mathbb{R}^{N^2}} \left\{ \frac{1}{2} \big\| \boldsymbol{\Phi} \boldsymbol{w} - \boldsymbol{u} \big\|_{2}^{2}  + \frac{\lambda}{2} \| \boldsymbol{w} \|_{2}^{2}\right\}.
\end{equation}
Thus, the regularization parameter $\lambda$ can be interpreted as the noise variance scaling in the Gaussian model. This equivalence between ridge regression and maximum {\it a posteriori} estimation under a Gaussian prior is well established in the statistical learning and machine learning literature~\citep{bishop2006pattern, hastie2009elements, wang2022gaussian}. The first term enforces fidelity to the observed data, in particular, controlling the squared bias, while the second penalizes large coefficients to reduce the variance, thus promoting smoothness and numerical stability.

The objective function in~\eqref{MINIMIZATION_PROBLEM} is strictly convex for any $\lambda > 0$, yielding the unique minimizer
\begin{equation}
    \label{SOLUTION_LEAST_SQUARES}
    \widehat{\boldsymbol{w}} = \big(\boldsymbol{\Phi}^{T}\boldsymbol{\Phi} + \lambda \boldsymbol{I}\big)^{-1} \boldsymbol{\Phi}^{T}\boldsymbol{u}.
\end{equation}
The regularized system in \eqref{SOLUTION_LEAST_SQUARES} can be solved using several matrix factorizations. In principle, the Cholesky decomposition is computationally efficient and numerical stable when $\boldsymbol{\Phi}^{T}\boldsymbol{\Phi} + \lambda \boldsymbol{I}$ is well-conditioned, as it exploits the symmetry and positive definiteness of the system. However, for poorly conditioned or highly correlated basis matrices, more numerically robust alternatives such as QR or singular-value decomposition (SVD) are often preferred~\cite{golub2013matrix}. We use the SVD decomposition in this paper.

While the regularized parameter $\lambda$ in \eqref{MINIMIZATION_PROBLEM} can be interpreted as the noise variance scaling in the Gaussian prior, its optimal value is generally unknown and has a strong influence on both bias and variance. Choosing $\lambda$ manually or via a coarse grid search can computationally expensive and problem-dependent~\cite{li2023automatic,meanti2022efficient}. To address this, we determine $\lambda$ automatically using the Generalized Cross-Validation (GCV), a statistically motivated, data-driven approach for selecting regularization parameters in linear inverse problems~\cite{golub1979generalized, hastie2009elements, lukas2006robust}. The idea is to estimate the predictive risk that would be obtained by leave-one-out cross-validation, without explicitly refitting the model for each data point. Denoting by 
\begin{equation*}
    \widehat{\boldsymbol{w}}_{\lambda} = \big(\boldsymbol{\Phi}^{T}\boldsymbol{\Phi} + \lambda \boldsymbol{I}\big)^{-1} \boldsymbol{\Phi}^{T}\boldsymbol{u}
\end{equation*}
the regularized estimator and by 
\begin{equation*}
    \boldsymbol{H}_{\lambda} = \boldsymbol{\Phi}\big(\boldsymbol{\Phi}^{T}\boldsymbol{\Phi} + \lambda \boldsymbol{I}\big)^{-1}\boldsymbol{\Phi}^{T}
\end{equation*}
the corresponding hat (or influence) matrix, the GCV score is defined as
\begin{equation*}
    \mathrm{GCV}(\lambda) = \frac{\| \boldsymbol{u} - \boldsymbol{\Phi}\widehat{\boldsymbol{w}}_{\lambda} \|_{2}^{2}}{\left(n - \mathrm{tr}(\boldsymbol{H}_{\lambda}) \right)^{2}}.
\end{equation*}
The denominator $\left( n - \mathrm{tr}(\boldsymbol{H}_{\lambda}) \right)^{2}$ corrects for model complexity, where $\mathrm{tr}(\boldsymbol{H}_{\lambda})$ measures the effective degree freedom of the model. Minimizing this quantity with respective to $\lambda$ provides a data-driven compromise between fidelity and smoothness, automatically balancing bias and variance. With an SVD decomposition of $\boldsymbol{\Phi}$ at hand, $\boldsymbol{H}_{\lambda}$ can be easily computed for any $\lambda$ using the former factorization. Once $\lambda^{\ast} = \argmin_{\lambda>0} \mathrm{GCV}(\lambda)$ is identified, the coefficient are computed in the same fashion, providing a stable and reproducible solution.

Given the fitted coefficients \(\bar{\boldsymbol{w}}\), the solution at any time \(t > 0\) and spatial location \((x, y)\) is predicted as
\begin{equation*}
    \label{PREDICTED_EQUATION}
    u(x, y, t) = \sum_{j=1}^{N} \sum_{k=1}^{N} \bar{w}_{j,k} \sin{\left( \frac{j\pi x}{L_{1}} \right)} \sin{\left( \frac{k\pi x}{L_{2}} \right)} \cos{\left( \omega_{j,k}\,t \right)}
\end{equation*}
which can be vectorized as 
\[
    u(x,y,t) = \boldsymbol{\Phi}(x,y)\, \text{diag}\left( \cos{\left( \omega_{j,k}\,t \right)} \right) \bar{\boldsymbol{w}}
\]
where \(\boldsymbol{\Phi}\colon\Omega\to\mathbb R^{1\times N^2}\) contains the spatial basis evaluated at \((x,y)\).

\subsection{Finite Element Method with Crank--Nicolson intergrator} Using the classical weak formulation of second-order hyperbolic equations~\citep{ern2004theory, larson2013finite, renardy2004introduction}, we represent the wave equation in the Hilbert space $V = H_{0}^{1}(\Omega)$ endowed with the bilinear form $a(u, v) = \int_{\Omega} \nabla u \cdot \nabla v \, \mathrm{d}x\mathrm d y $ for all $u, v \in V$. On the strength of Poincar\'{e}-Friedrichs inequality, the latter constitutes a norm equivalent with the standard Sobolev $H^1$-norm on $V$. By letting $H = L^{2}(\Omega)$, we obtain the Gelfand triple $(V, H, V')$, and the weak formulation seeks $u\in C^{2}\left([0, T], V'\right) \cap C^{1}\left([0, T], H\right) \cap C^{0}\left([0, T], V\right)$ such that for all $v \in V$ 
\begin{equation}
    \label{WEAK_FORMULATION_EQUATION}
    \left\langle\partial_{tt} u(t), v \right\rangle_{H} + c^2 a(u(t), v) = 0
\end{equation}
for $t \in [0, T]$ with initial data $u(0) = u_{0}$, $u_{t}(0) = v_{0}$. This formulation implies the energy conservation identity
\begin{equation*}
    \frac{\mathrm{d}E(t)}{\mathrm{d}t} = 0 \text{ or } E(t) \equiv E(0),
\end{equation*}
where
\begin{equation*}
    E(t) = \frac{1}{2}\big(\| u_{t}(t) \|_{H}^{2} + c^{2} \| \nabla u(t) \|_{H}^{2}\big)
\end{equation*}
is the energy of the homogeneous system~\citep{evans2022partial,renardy2004introduction}.

Let $\mathcal{T}_{h}$ be a conforming triangulation of $\Omega$ with the mesh size $h$, and let $V_{h} \subset V$ be the finite element space defined as 
\begin{equation*}
    V_{h} = \left\{ v_{h} \in V : v_{h}|_{K} \in \mathcal{P}_{1}(K) \text{ for each } K \in \mathcal{T}_{h} \right\},
\end{equation*}
where $\mathcal{P}_{1}(K)$ denotes the space of polynomials on $K$ of up to degree 1. We approximate $u(\cdot, t)$ by 
\begin{equation}
    \label{LINEAR_FINITE_ELEMENT_EQUATION}
    u_{h}(x, t) = \sum_{i=1}^{N}U_{i}(t) \varphi_{i}(x),
\end{equation}
where $\{\varphi_{i}\}_{i = 1}^{N}$ is the nodal basis of $V_{h}$. Substituting~\eqref{LINEAR_FINITE_ELEMENT_EQUATION} into the weak form \eqref{WEAK_FORMULATION_EQUATION} and taking $v_{h} = \varphi_{j}$ yields the semi-discrete finite element system
\begin{equation}
    \label{SEMI_DISCETE_SYSTEM}
    \boldsymbol{M} \ddot{\boldsymbol{U}}(t) + c^{2} \boldsymbol{K}\boldsymbol{U}(t) = 0,
\end{equation}
where $\boldsymbol{U} = \left(U_{1}, \dots, U_{N} \right)^{T} \in \mathbb{R}^{N} $ is the vector of nodal values, $\boldsymbol{M}$ is the constant mass matrix with the entries
\begin{equation*}
    M_{i, j} = \int_{\Omega} \varphi_{i}(x,y) \varphi_{j}(x, y) \,\mathrm{d}x\, \mathrm{d}y
\end{equation*}
and $\boldsymbol{K}$ is the constant stiffness matrix with  entries
\begin{equation*}
    K_{i, j} = \int_{\Omega} \nabla \varphi_{i}(x, y) \cdot \nabla \varphi_j(x, y) \, \mathrm{d}x\,\mathrm{d}y.
\end{equation*}
For each triangular element $T_{e}$ of area $A_e$, the element of the mass and stiffness matrix are defined as
\begin{equation*}
     \boldsymbol{M}_{e} = \frac{A_{e}}{12} \begin{pmatrix}
        2 & 1 & 1 \\
        1 & 2 & 1 \\
        1 & 1 & 2
    \end{pmatrix}, \quad \boldsymbol{K}_{e, (i,j)} = \frac{1}{4A^{e}}(b_{i}\,b{j} + c_{i}\,c_{j}),
\end{equation*}
respectively, where \(b_{i} = y_{i+1} - y_{i-1}\) and \( c_{i} = x_{i-1} - x_{i+1}\) (with cyclic indexing)~\citep{ciarlet2002finite, larson2013finite}. The global matrices are assembled by summing contributions from all elements.

Let $t_{n} = n (\Delta t)$ with the uniform stepsize $\Delta t := \frac{T}{N_t} $ for $n = 0, 1, \dots, N_{t}$, and denote $\boldsymbol{U}^{n} \approx \boldsymbol{U}(t_{n})$. Applying the Crank--Nicolson scheme to the semi-discrete system \eqref{SEMI_DISCETE_SYSTEM} gives
\begin{equation*}
    \boldsymbol{M} \dfrac{\boldsymbol{U}^{n+1} - 2\boldsymbol{U}^{n} + \boldsymbol{U}^{n-1}}{(\Delta t)^{2} } + c^{2} \boldsymbol{K} \dfrac{\boldsymbol{U}^{n+1} - \boldsymbol{U}^{n-1}}{2} = 0.
\end{equation*}
Introducing $\alpha = \frac{c^{2} (\Delta t)^{2}}{2}$, we can rewrite the iterative update as
\begin{equation*}
    \left( \boldsymbol{M} + \alpha \boldsymbol{K} \right) \boldsymbol{U}^{n+1} = \left( 2\boldsymbol{M} - \alpha \boldsymbol{K} \right)\boldsymbol{U}^{n} - \boldsymbol{M} \boldsymbol{U}^{n-1}.
\end{equation*}

The Crank--Nicolson scheme is unconditionally stable for the homogeneous wave equation, allowing larger time steps compared to explicit methods without stability restrictions. Also, on regular solutions, namely, $u \in C^{3}\big([0, T], L^{2}(\Omega)\big) \cap C\big([0, T], H^{2}(\Omega)\big)$ and regular meshes, this method achieves second-order accuracy in both space and time, providing an \(\mathcal{O}(h^{2} + (\Delta t)^2)\) in convergence in $C^{0}([0, T], L^{2}(\Omega)\big)$. Note that since the left-hand side of the matrix \(M + \alpha K \) remains constant throughout the implementation, it can be factorized only once at the beginning, making each time step computationally efficient.

\section{Numerical Experiments}\label{sec:experiments}
We benchmark our proposed method against the classical Crank--Nicolson finite element method (CN-FEM) on two representative test problems. Our experiments are designed to ensure a fair comparison by matching for model complexity based on degrees of freedom (DoF) generating high-accuracy reference solutions, and reporting both accuracy and efficiency metrics.

Because B-EPGP and CN-FEM have fundamentally different computational structures, we compare them under matched degrees of freedom (DoF). For B-EPGP, we define the (effective) DoF as
\begin{equation*}
    \text{DoF}_{\text{EP}} = \text{tr}\big[\boldsymbol{\Phi}\big( \boldsymbol{\Phi}^{T} \boldsymbol{\Phi} + \lambda \boldsymbol{I} \big)^{-1}\boldsymbol{\Phi}^{T} \big]
\end{equation*}
consistent with the formulation of effective degrees of freedom in linear regression, smoothing, and regularization, where the degrees of freedom are expressed as the trace of the associated smoothing or hat matrix~\cite{hastie2017generalized, janson2015effective, wahba1990spline}. This formulation quantifies the model's flexibility by measuring the sensitivity of the fitted values to the observed data and incorporates the influence of the regularization parameter $\lambda$. For CN-FEM with homogeneous Dirichlet boundary conditions, the DoF depends on both spatial and temporal discretization, which is given by
\begin{equation*}
    \text{DoF}_{\text{CN}} = (n-1)^2 \times N_{t}
\end{equation*}
where $n$ denotes the number of spatial points per coordinate direction, yielding $(n-1)^{2}$ interior nodes, and $N_t = T/\Delta t + 1 $ represents the number of time steps. This definition follows the usual finite element interpretation of global degrees of freedom as the number of independent nodal unknowns remaining after enforcing boundary constraints~\cite{langtangen2017solving}.

To ensure a fair comparison between the two solvers, we select $(n, \Delta t)$ such that $\text{DoF}_{\text{CN}} \approx \text{DoF}_{\text{EP}}$. To maintain consistency between spatial and temporal discretizations, we couple the time step to the spatial resolution by setting $\Delta t = T/n$. This choice yields $N_t = n + 1$, eliminating the explicit dependence on $T$ in the degree of freedom relation. Substituting into the Crank--Nicolson formulation gives the cubic equation 
\begin{equation*}
    (n-1)^{2}(n+1) = \mathrm{DoF}_{\mathrm{EP}}
\end{equation*}
which is solved for the real positive root to obtain $n$, the corresponding $\Delta t$ then follows directly. Comparing computational methods at matched degrees of freedom provides a consistent basis for evaluating accuracy and cost. A similar approach has been adopted by~\cite{vondvrejc2020energy}, who compared the Fourier–Galerkin and finite element methods under equivalent numerical resolution.
High accuracy reference solutions were computed using the CN-FEM on fine grids (\(400 \times 400\)) with smaller time steps smaller that the spatial grid spacing ($\Delta t < 1/n$), ensuring that the temporal discretization error remains negligible compared to spatial discretization error. These solutions serve as ground truth for error evaluation.

For sufficiently smooth solutions of the linear wave equation, piecewise linear FEM in space combined with the Crank--Nicolson scheme in time is second-order accurate in both space and time. The global discretization error satisfies an {\it a priori} bound of the form
\begin{equation*}
    \| u_{h} - u_{\text{ref}} \|_{L^{\infty}(0, T; L^{2}(\Omega))} = \max_{0 \leq t_{i} \leq T} \|u(\cdot, t_i) - u_{h}^{i}\|_{L^{2}(\Omega)} \le C\big(h^2 + (\Delta t)^{2}\big),
\end{equation*}
where $C$ is independent of $h$ and $\Delta t$~\cite{baker1976error, georgoulis2013posteriori,larson2013finite}. On the unit square, $h \approx 1 / n$, and under our DoF-matching rule $\Delta t = 1/n$, the bounds scales as $h^2 + (\Delta t)^2 = 2/n^{2}$. Let $n_{\mathrm{ref}}$ denote the fine reference grid, while $n \ll n_{\mathrm{ref}}$ corresponds to coarser grids used in the experiments. Then asymptotic relative errors becomes
\begin{equation*}
    \frac{h^{2} + (\Delta t)^2}{h_{\mathrm{ref}}^{2} + (\Delta t_\mathrm{ref})^2} = \left( \frac{n_\mathrm{ref}}{n}\right)^{2}
\end{equation*}
indicating the reference solution is $(n_\mathrm{ref}/n)^{2}$ times more accurate in the asymptotic regime. As an illustration, with $n_\mathrm{ref} = 400$ and representative experimental grid $n = 28$, the ratio is $(400/28)^2 \approx 2.04 \times 10^2$ meaning the reference solution's theoretical error bound is roughly two orders of magnitude smaller than that of the coarser DoF-matched run. This justifies treating the fine-grid CN-FEM solution as a numerically exact reference in subsequent error analyses.
To ensure consistent quantitative comparison between the CN-FEM and B-EPGP, we evaluate the discrepancy between the numerical and reference solution using space-time $L^{2}$ error norm.

Let $u_{\text{ref}}(x, y, t)$ denote the high-resolution CN-FEM solution defined on the continuous space-time $L^{2}$-norm~\citep{ern2004theory, quarteroni1994numerical} as 
\begin{equation}
    \label{SPACE_TIME_ERROR}
    \| u_{h} - u_{\text{ref}} \|_{L^{2}(0, T; L^{2}(\Omega))}^{2} = \int_{0}^{T}\int_{\Omega} \big|u_{h}(x, y, t) - u_{\text{ref}}(x,y,t)\big|^{2} \, \mathrm{d}(x, y) \,\mathrm{d}t.
\end{equation}
This metric naturally arises in the analysis of time-dependent problems and provides a consistent ``isotropic'' measure of the overall space-time accuracy.
Following the standard finite element practice~\cite{ern2004theory, larson2013finite}, the spatial integration is approximated via Gaussian quadrature on each triangle $T_{e} \in \Omega_h$. Let $(x_{e,q}, y_{e,q})$ and $w_{q}$ denote quadrature points and weights, and $|T_{e}|$ the element area. Then for any sufficiently smooth $f(x, y)$,
\begin{equation}
    \label{GAUSSIAN_QUAD}
    \int_{\Omega} f(x, y) \mathrm{d}(x, y) \approx \sum_{e \in \mathcal{T}_{h}} |T_{e}| \sum_{q=1}^{4} w_{q} f(x_{e,q}, y_{e,q}).
\end{equation}
This quadrature rule is exact for all polynomials up to a total degree three and therefore achieves fourth-order accuracy in space for smooth integrands~\citep{dunavant1985high}. Both CN-FEM and B-EPGP solutions, as well as the reference solution, are evaluated at the same quadrature points on the CN mesh using bilinear interpolation from a finer reference grid. 
Temporal integration was performed using a uniform reference time discretization and the composite Simpson's $1/3$ rule~\citep{suli2003numerical}, which is exact for cubic functions and yields fourth-order accuracy in time. This combination ensures that both spatial and temporal quadratures have comparable high-order accuracy, allowing the reported space-time error to reflect the true discretization performance of the solvers rather than the numerical integration scheme. 
The reference solution $u_{\text{ref}}(\cdot, t)$ at any arbitrary time $t$ is obtained by linear interpolation. 
 Let $E^{2}_{n} = \|u_{h}(\cdot, t_{n}) - u_{\text{ref}}(\cdot, t_{n})\|_{L^{2}(\Omega)}^{2}$. Then the discrete approximation of the space-time norm in \eqref{SPACE_TIME_ERROR} is given by
\begin{equation}
    \label{ERROR_EQN}
    \| u_{h} - u_{\text{ref}} \|^{2}_{L^{2}(0, T; L^{2}(\Omega))} \approx \frac{\Delta t}{3} \Big(\frac{1}{2} E_{0}^{2} + 4\sum_{k=1}^{N_t/2}E_{2k-1}^{2} + 
    2\sum_{k=1}^{N_t/2-1}E_{2k}^{2} + E_{N_t}^{2} \Big).
\end{equation}
This ensures the computed space-time error accurately reflects the solver performance.

To obtain a dimensionless accuracy indicator independent of the reference amplitude, we define the relative space-time $L^{2}$-error by
\begin{equation*}
    \epsilon_\mathrm{rel} = \dfrac{\| u_{h} - u_{\text{ref}} \|_{L^{2}((0, T), L^{2}(\Omega))}}{\| u_{\text{ref}} \|_{L^{2}((0, T), L^{2}(\Omega))}}.
\end{equation*}
The relative error measures the normalized $L^2$ distance between the numerical and reference solutions~\cite{ hughes2012finite,larson2013finite}. A smaller value indicates that the numerical approximation is closer to the reference solution in the mean square sense.

We consider two representative initial conditions that highlight different aspects of wave propagation. The first one is a smooth polynomial profile
\begin{equation*}
    \label{POLYNOMIAL_IC}
    u(x, y, 0) = x(1-x)y(1-y),
\end{equation*}
which vanishes on the boundary of the domain and, therefore, naturally satisfies the Dirichlet boundary conditions. The example serves as a well-behaved test case for assessing accuracy in the smooth setting.

The second initial condition employs a smooth, compactly supported (non-normalized) \emph{mollifier} of Friedrichs type \citep{evans2022partial,hormander2007analysis,renardy2004introduction} defined by
\begin{equation*}
    \label{MOLLIFIER_IC}
    u(x, y, 0) = 
    \begin{cases}
        \exp\!\left(-\dfrac{R^{2}}{R^{2} - (x-x_{0})^{2} + (y-y_{0})^{2}}\right) 
            & \text{if } (x-x_{0})^{2} + (y-y_{0})^{2} < R^{2}, \\[10pt]
        0   & \text{otherwise}.
    \end{cases}
\end{equation*}
Here $(x_{0}, y_{0}) = (0.3, 0.7)$ denotes the center of the bell-shaped ``bump,'' and $R = 0.24$ represents the radius of compact support beyond which the function vanishes. These are the parameter values used in the numerical experiments. The mollifier satisfies homogeneous boundary conditions by vanishing smoothly and to all orders at the boundary of its support.

\section{Results}\label{sec:results}

We evaluate the proposed B-EPGP surrogate against the CN-FEM for the two-dimensional wave equation on the unit square domain $\Omega=(0,1)^{2}$ with the wave speed $c=1.0$, homogeneous Dirichlet boundary conditions, and terminal time $T=1.0$. Table~\ref{tab:l2_space_time} shows the space--time $L^2$-errors and corresponding relative errors computed with respect to the high-resolution reference solution. All configurations were chosen to achieve comparable degrees of freedom (DoF), with the CN-FEM discretization adjusted to approximate the effective DoF of the B-EPGP model. Table~\ref{tab:linf_time} further contains the maximum-in-time spatial errors $L^{\infty}(0,T;L^{2}(\Omega))$, providing a complementary measure of temporal accuracy.

\begin{table}[h]
\centering
\caption{Space--time $L^{2}$- and relative errors for CN-FEM and B-EPGP.}
\begin{tabular}{lccccc}
\toprule
Initial Condition
& CN
& B-EPGP
& CN Relative
& B-EPGP Relative
& Improvement \\
& Error & Error & Error & Error \\
\midrule
Polynomial  & $7.70\times10^{-3}$ & $2.44\times10^{-5}$ & 31.79\% & 0.10\% & $315.60\times$ \\
Mollifier & $4.43\times10^{-2}$ & $5.68\times10^{-4}$ & 75.31\% & 0.96\% & $78.08\times$ \\
\bottomrule
\end{tabular}
\label{tab:l2_space_time}
\end{table}

\begin{table}[h]
\centering
\caption{$L^{\infty}(0,T;L^{2}(\Omega))$- and relative errors for CN-FEM and B-EPGP.}
\begin{tabular}{lccccc}
\toprule
Initial Condition
& CN
& B-EPGP
& CN Relative
& B-EPGP Relative
& Improvement \\
& Error & Error & Error & Error \\
\midrule
Polynomial & $1.20\times10^{-2}$ & $3.62\times10^{-5}$ & $36.07\%$ & $0.11\%$ & $332.37\times$ \\
Mollifier  & $6.59\times10^{-2}$ & $1.02\times10^{-3}$ & $79.93\%$ & $1.24\%$ & $64.64\times$ \\
\bottomrule
\end{tabular}
\label{tab:linf_time}
\end{table}

The relative error provides a measure of normalized deviation from the reference solution.
As shown in Table~\ref{tab:l2_space_time}, B-EPGP reduces the space-time relative $L^{2}$-error from approximately $32\%$ to $0.1\%$ in the smooth polynomial case and from $75.31\%$ to $0.96\%$ in the mollifier case, representing improvements of two to three orders of magnitude in overall accuracy.
Consistent trends are observed in Table~\ref{tab:linf_time}, where the maximum-in-time spatial $L^{2}$-error decreases by roughly two orders of magnitude in both tests. These results confirm that the operator-informed basis of B-EPGP provides substantially higher fidelity than the implicit CN-FEM under matched DoF. 
It should be noted that the fine-grid CN-FEM reference is treated as the true solution for quantitative comparison. If the exact analytical solution were available, the true error of the B-EPGP surrogate would likely be even smaller, since the high-resolution reference itself contains a small residual discretization error. This demonstrates that the reported improvements are conservative.

For all B-EPGP experiments, we used $N = 40$ frequency modes per spatial direction and $n = 5{,}000$ sample points for basis construction. The spatial samples were generated using a Latin hypercube design (LHD), which ensures space-filling coverage of the domain while minimizing correlations between coordinates. This sampling strategy allows the B-EPGP surrogate to represent the spatial variability of the solution effectively with fewer evaluations than a uniform grid. The regularization parameter $\lambda$ was selected automatically using GCV, from which the effective DoF were computed.

\medskip

\paragraph{\textbf{Polynomial}.}
The optimal $\lambda^{\ast} = 1.00\times10^{-6}$ produced an effective DoF of $1{,}600$, while the matched CN-FEM discretization ($n=12$) contained $1{,}573$ DoF. At this complexity, the space--time $L^2$-error decreased from $7.70\times10^{-3}$ for CN-FEM to $2.44\times10^{-5}$ for B-EPGP, a $315.60\times$ improvement, with the relative error dropping from $31.79\%$ to $0.10\%$. The maximum-in-time spatial error was also  two orders of magnitude smaller for B-EPGP, decreasing from $1.20\times10^{-2}$ for CN-FEM to $3.62\times10^{-5}$, with the corresponding relative error reduced from $36.07\%$ to $0.11\%$. Although the total runtime includes the GCV search (1.441\,s), the fitting step after selecting the optimal $\lambda$ was completed almost instantaneously, while the CN-FEM solve required 0.040\,s.

\ifdefined\INCLUDEFIGS

\begin{figure}[h]
  \vskip 0.2in
  \centering
  \begin{subfigure}[b]{0.15\textwidth}
    \centering
    \includegraphics[width=\textwidth]{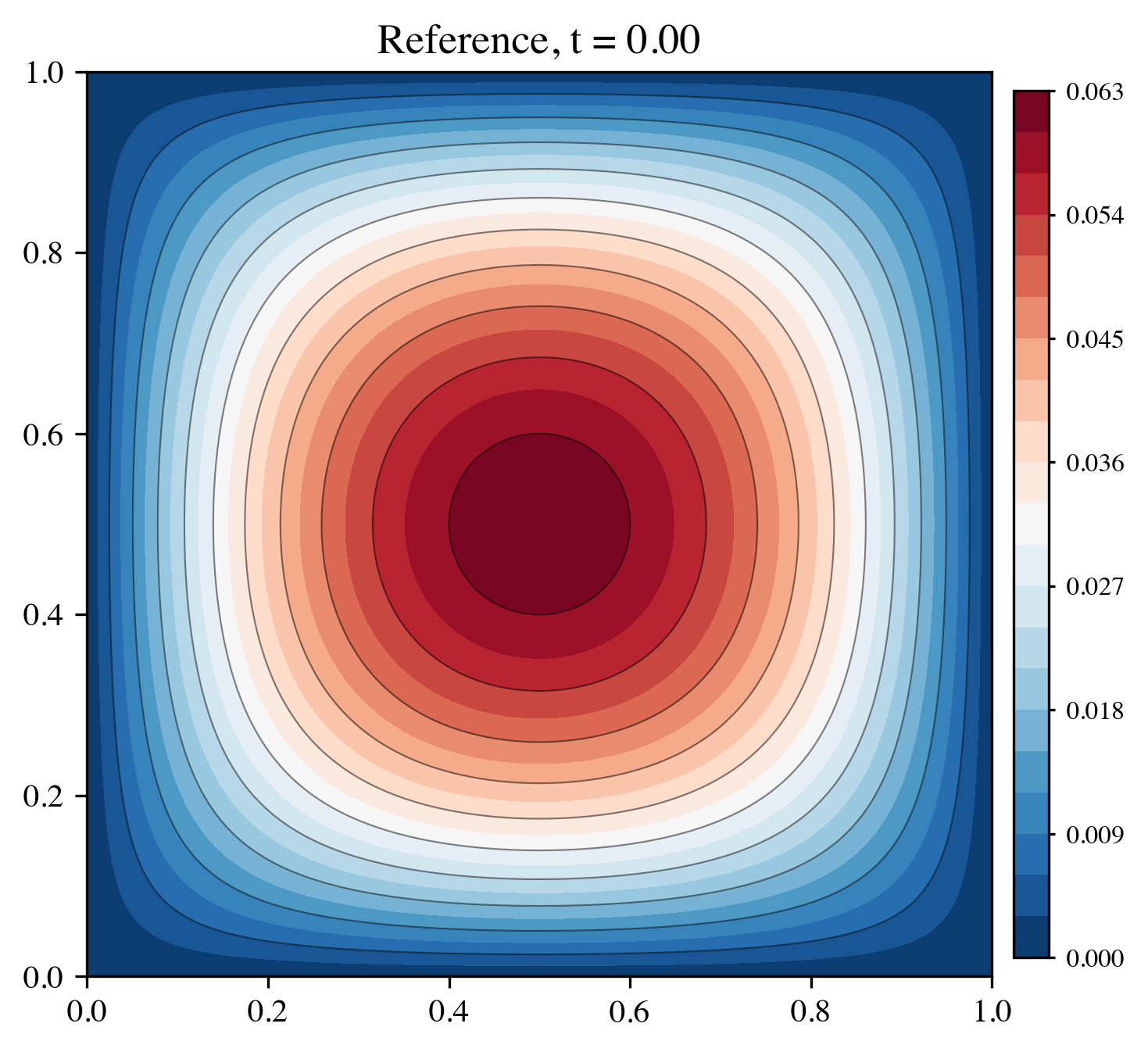}
    \caption{\tiny $t=0.0$,\\ Reference}
  \end{subfigure}
  \hfill
  \begin{subfigure}[b]{0.15\textwidth}
    \centering
    \includegraphics[width=\textwidth]{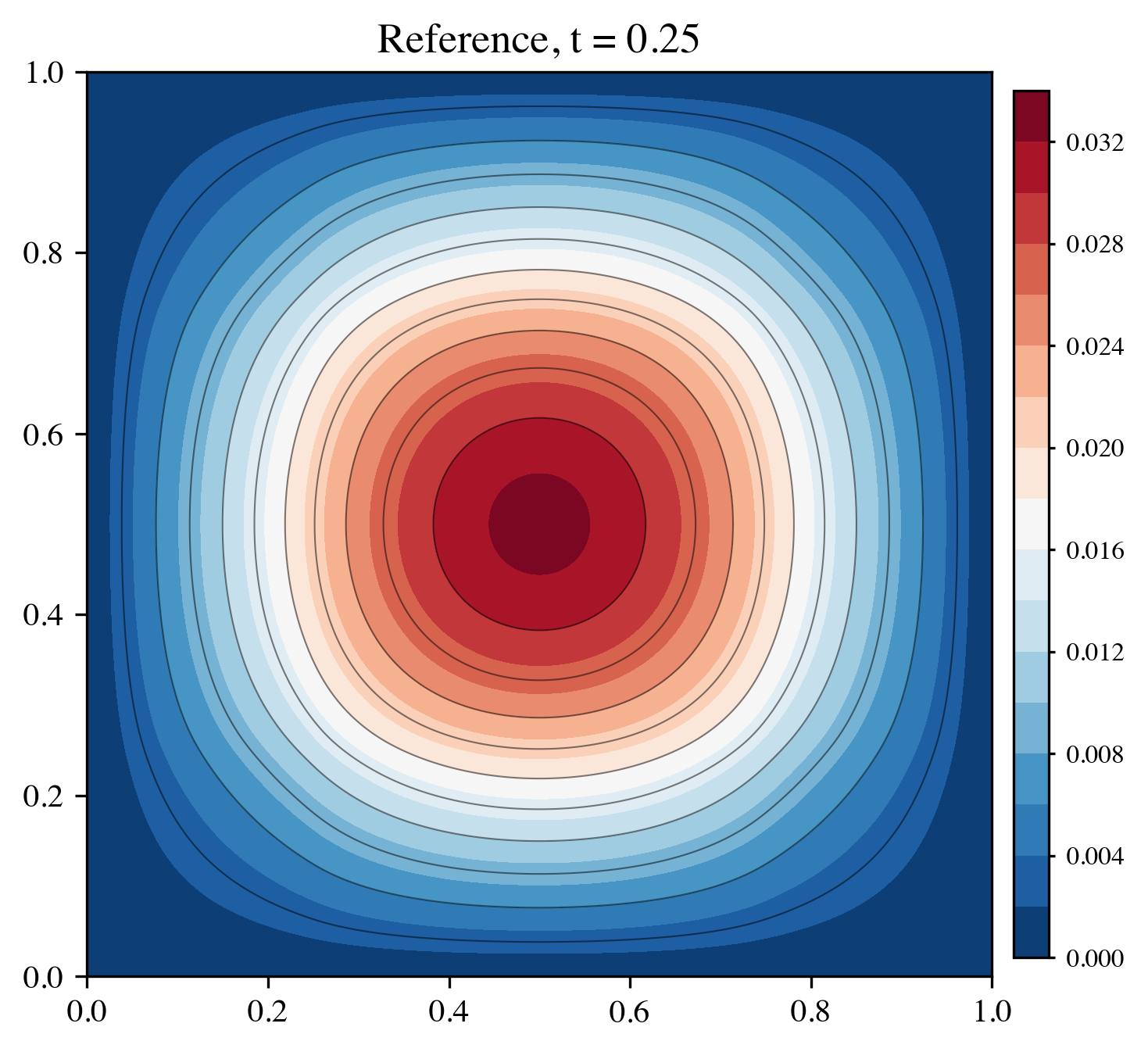}
    \caption{\tiny $t=0.25$,\\ Reference}
  \end{subfigure}
  \hfill
  \begin{subfigure}[b]{0.15\textwidth}
    \centering
    \includegraphics[width=\textwidth]{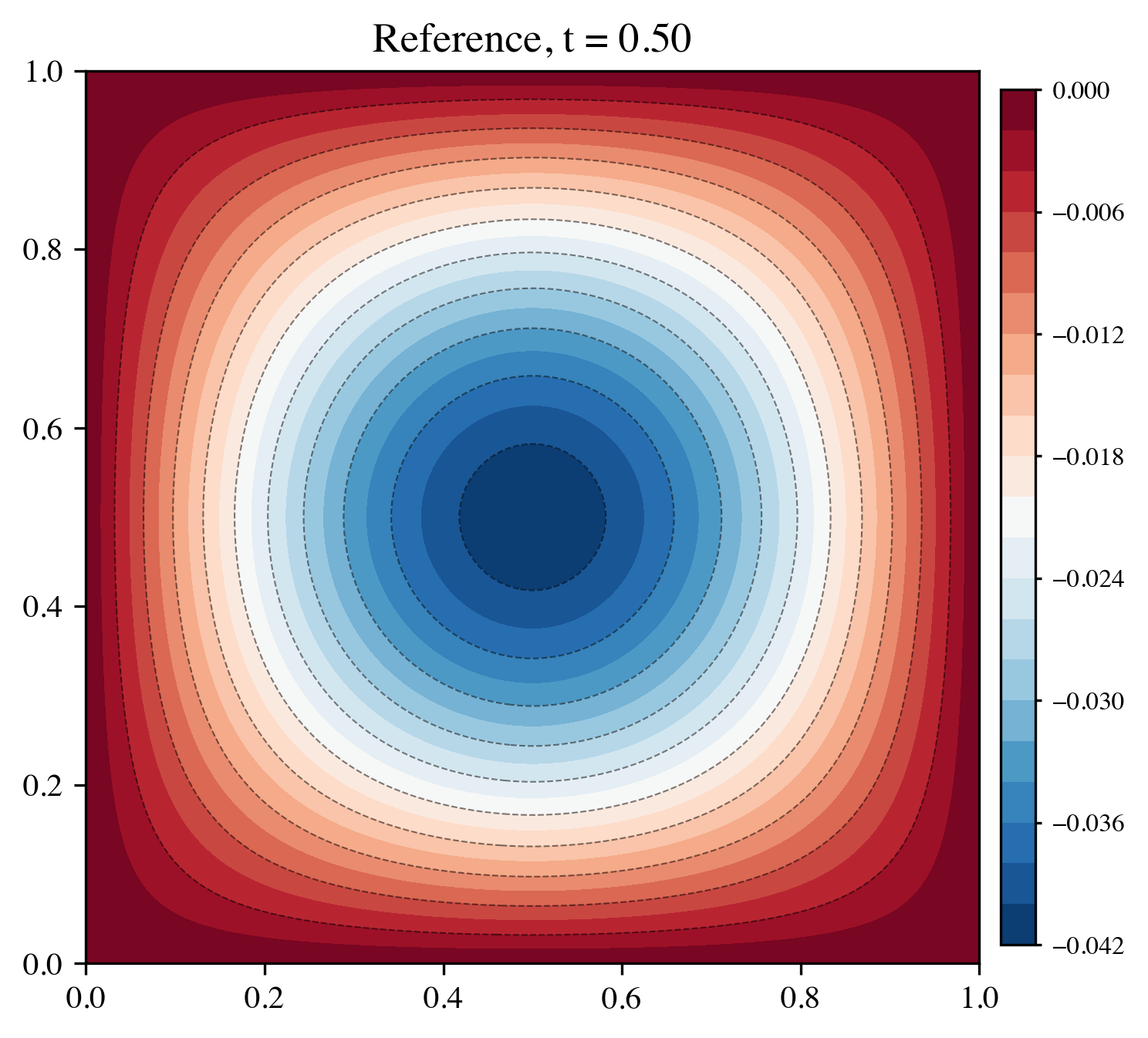}
    \caption{\tiny $t=0.50$,\\ Reference}
  \end{subfigure}
  \hfill
  \begin{subfigure}[b]{0.15\textwidth}
    \centering
    \includegraphics[width=\textwidth]{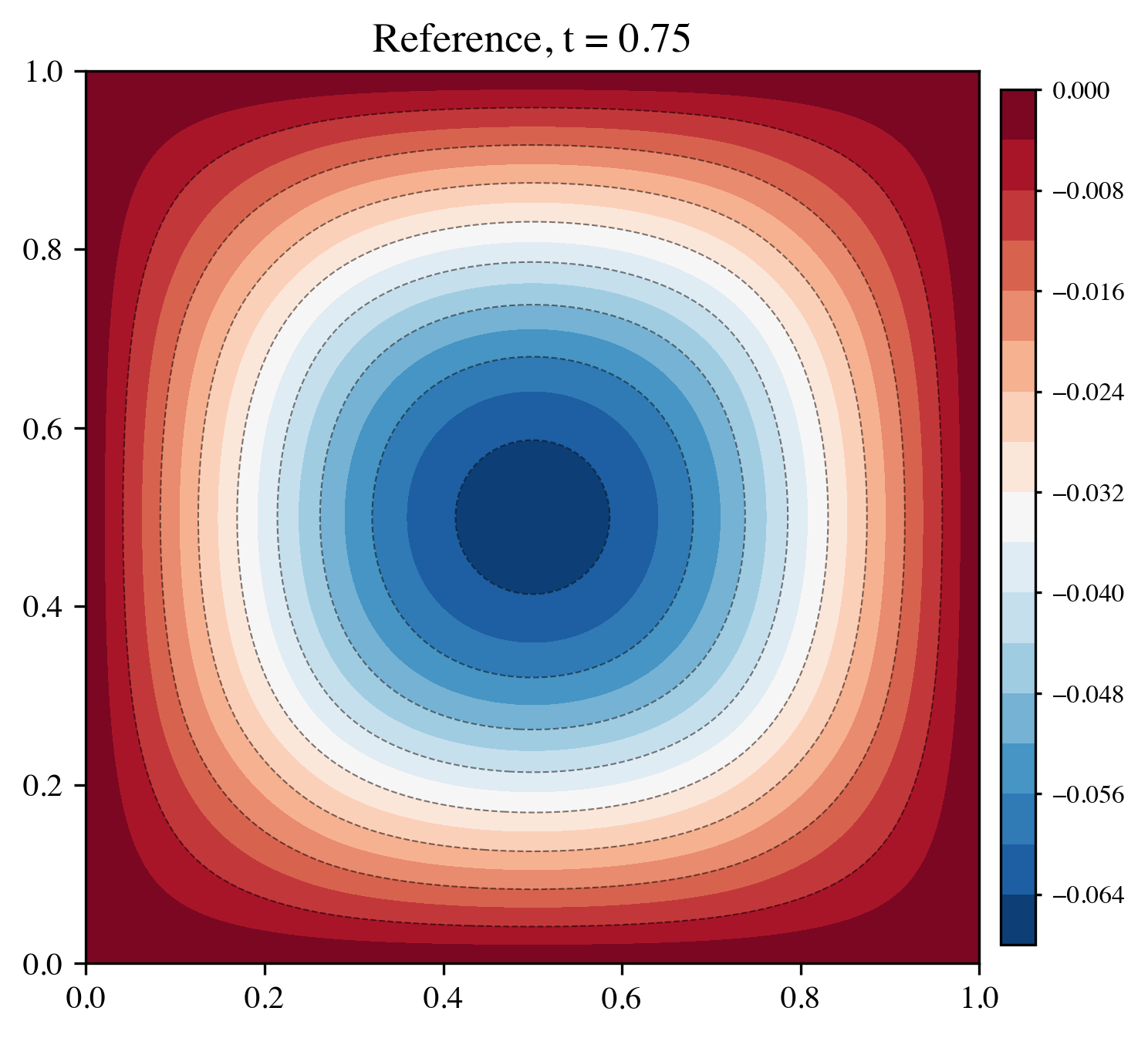}
    \caption{\tiny $t=0.75$,\\ Reference}
  \end{subfigure}
  \hfill
  \begin{subfigure}[b]{0.15\textwidth}
    \centering
    \includegraphics[width=\textwidth]{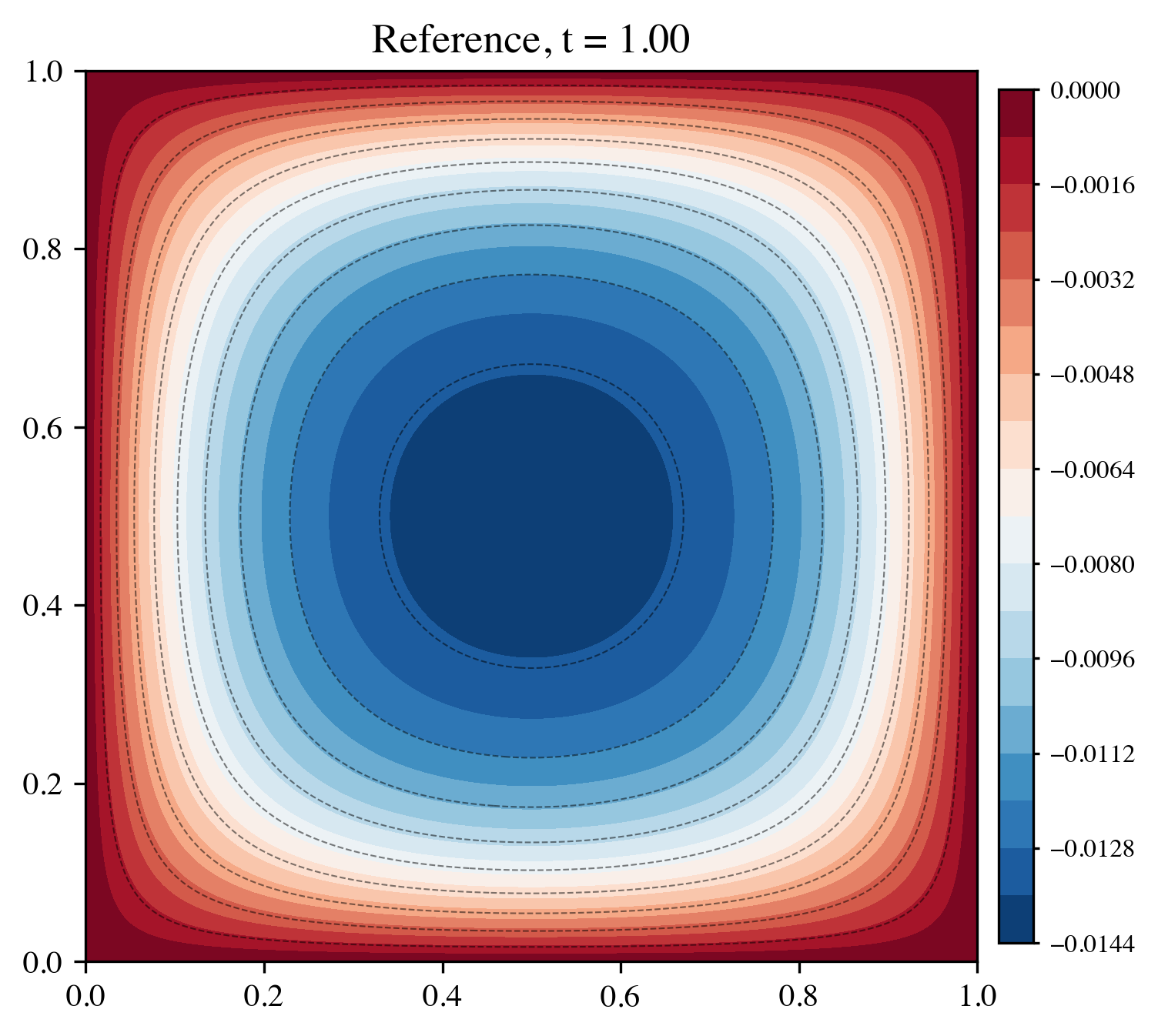}
    \caption{\tiny $t=1.0$,\\ Reference}
  \end{subfigure}
  \\
  \begin{subfigure}[b]{0.15\textwidth}
    \centering
    \includegraphics[width=\textwidth]{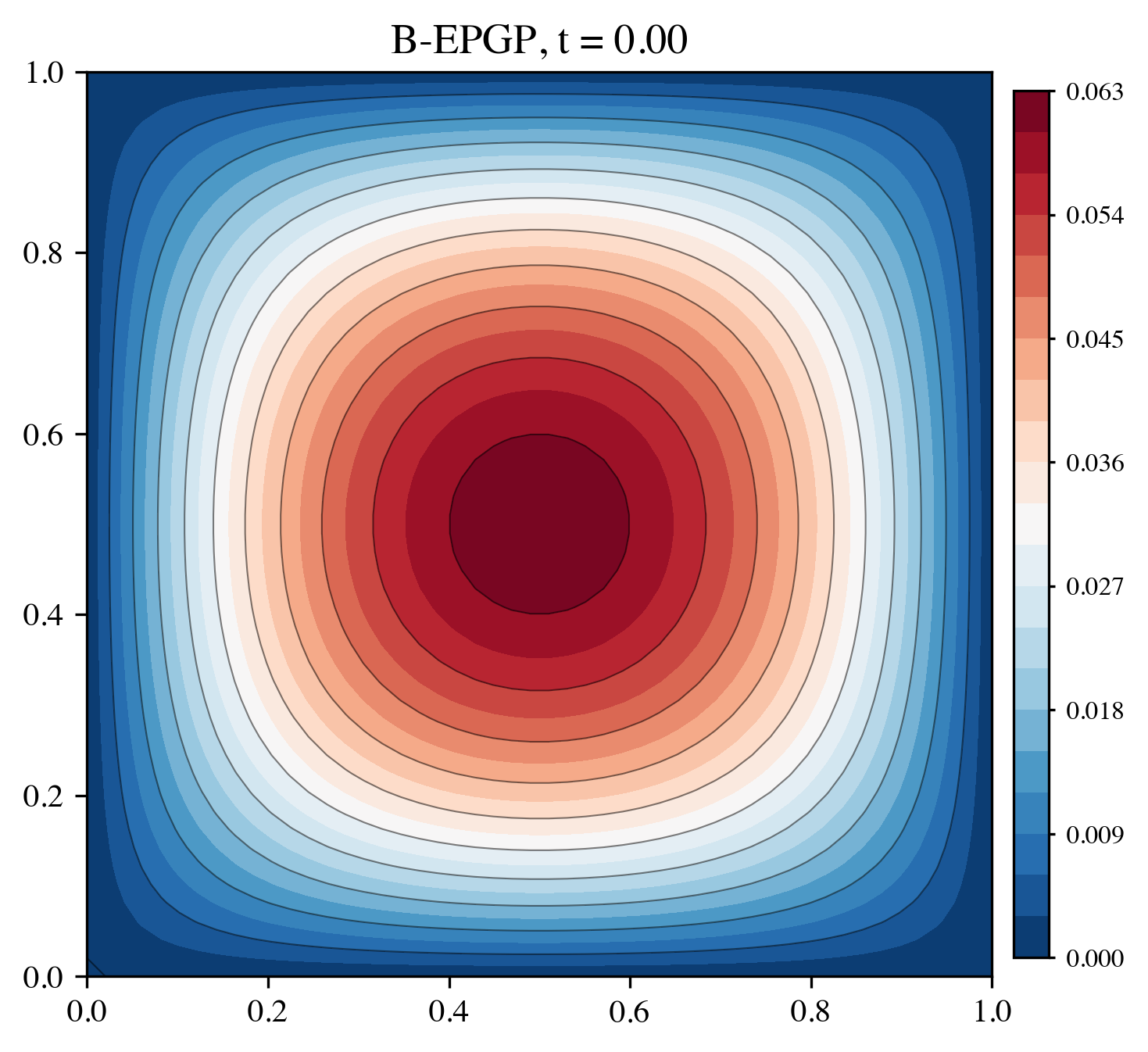}
    \caption{\tiny $t=0.0$,\\ B-EPGP}
  \end{subfigure}
  \hfill
  \begin{subfigure}[b]{0.15\textwidth}
    \centering
    \includegraphics[width=\textwidth]{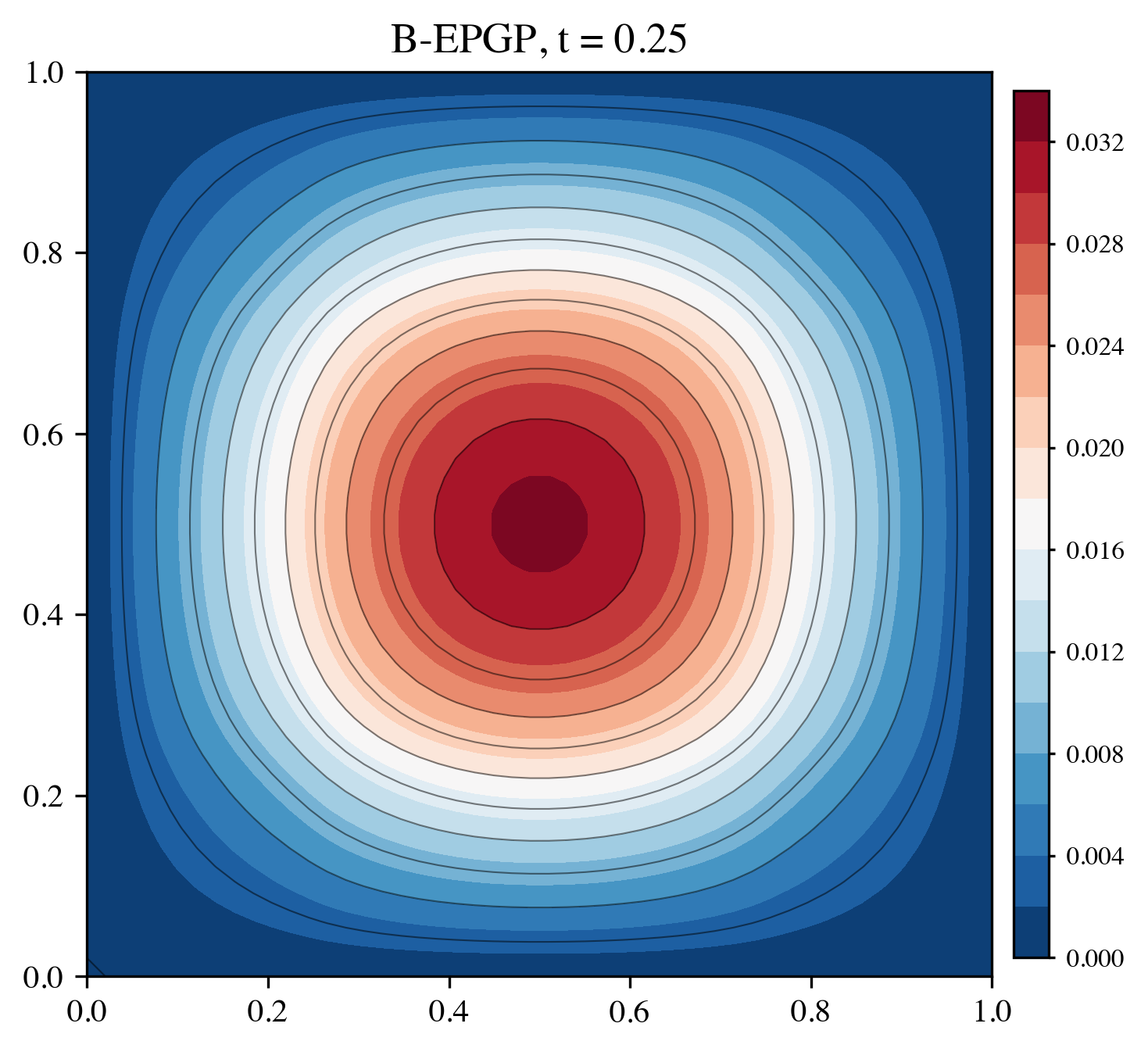}
    \caption{\tiny $t=0.25$,\\ B-EPGP}
  \end{subfigure}
  \hfill
  \begin{subfigure}[b]{0.15\textwidth}
    \centering
    \includegraphics[width=\textwidth]{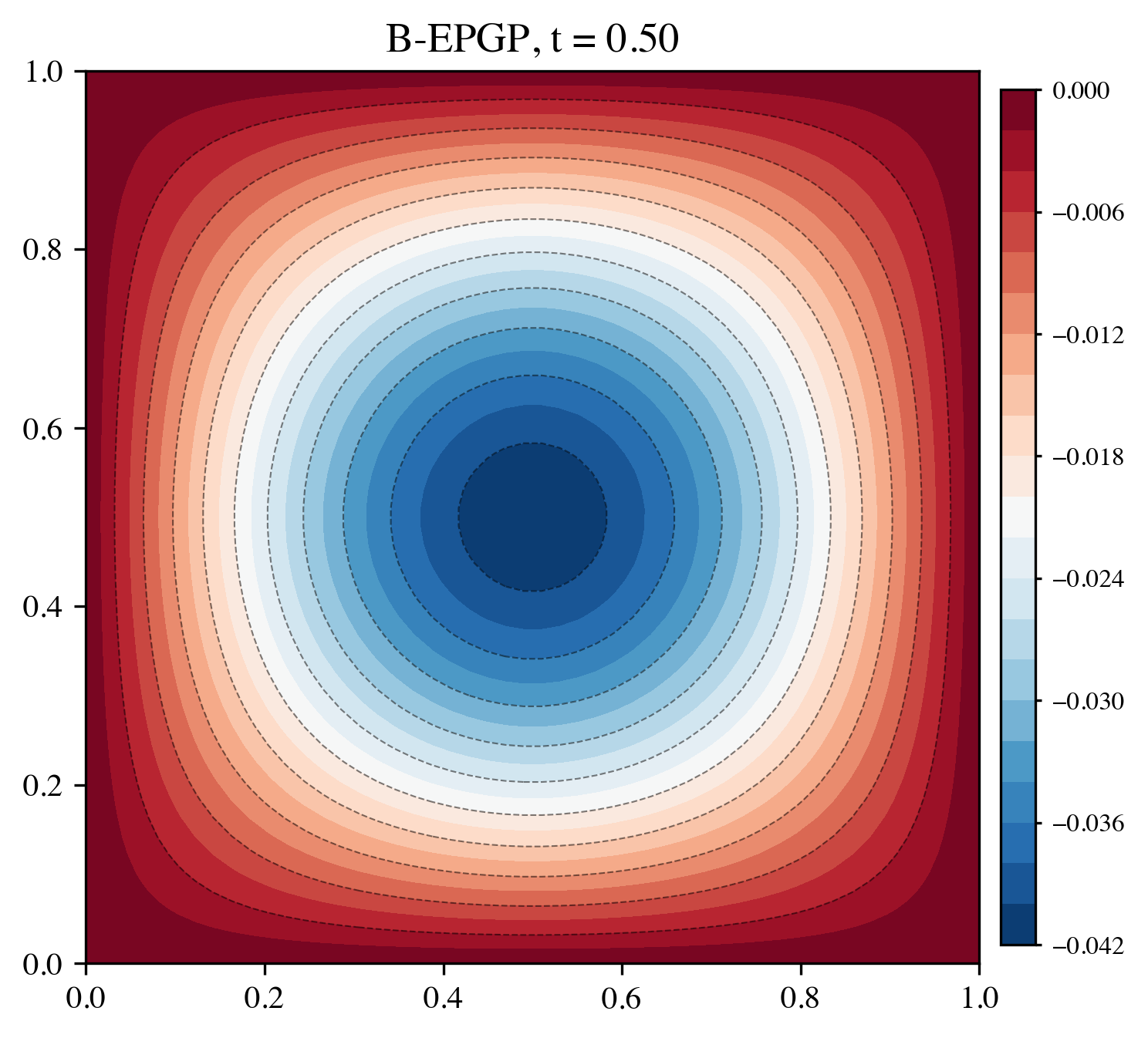}
    \caption{\tiny $t=0.50$,\\ B-EPGP}
  \end{subfigure}
  \hfill
  \begin{subfigure}[b]{0.15\textwidth}
    \centering
    \includegraphics[width=\textwidth]{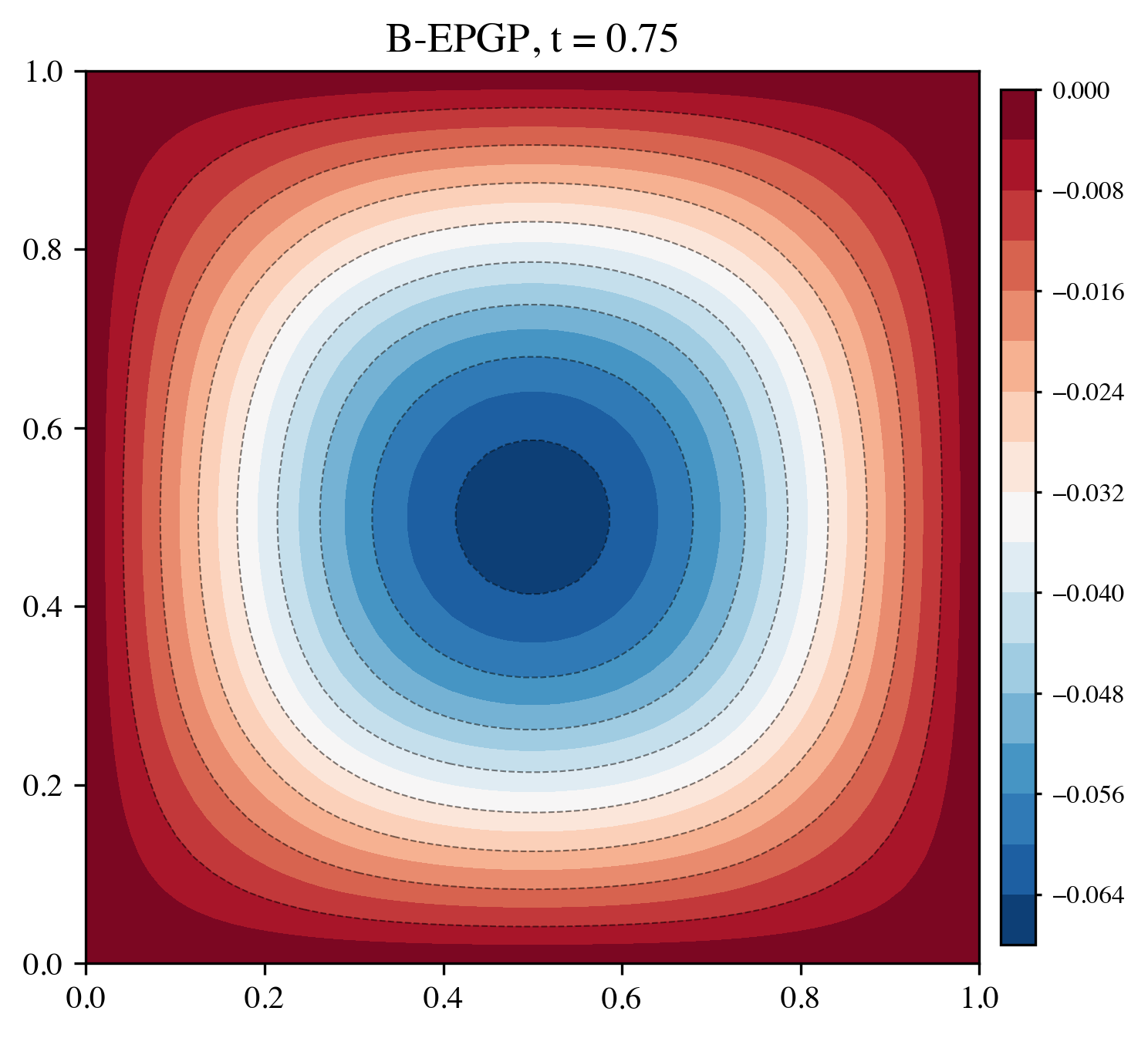}
    \caption{\tiny $t=0.75$,\\ B-EPGP}
  \end{subfigure}
  \hfill
  \begin{subfigure}[b]{0.15\textwidth}
    \centering
    \includegraphics[width=\textwidth]{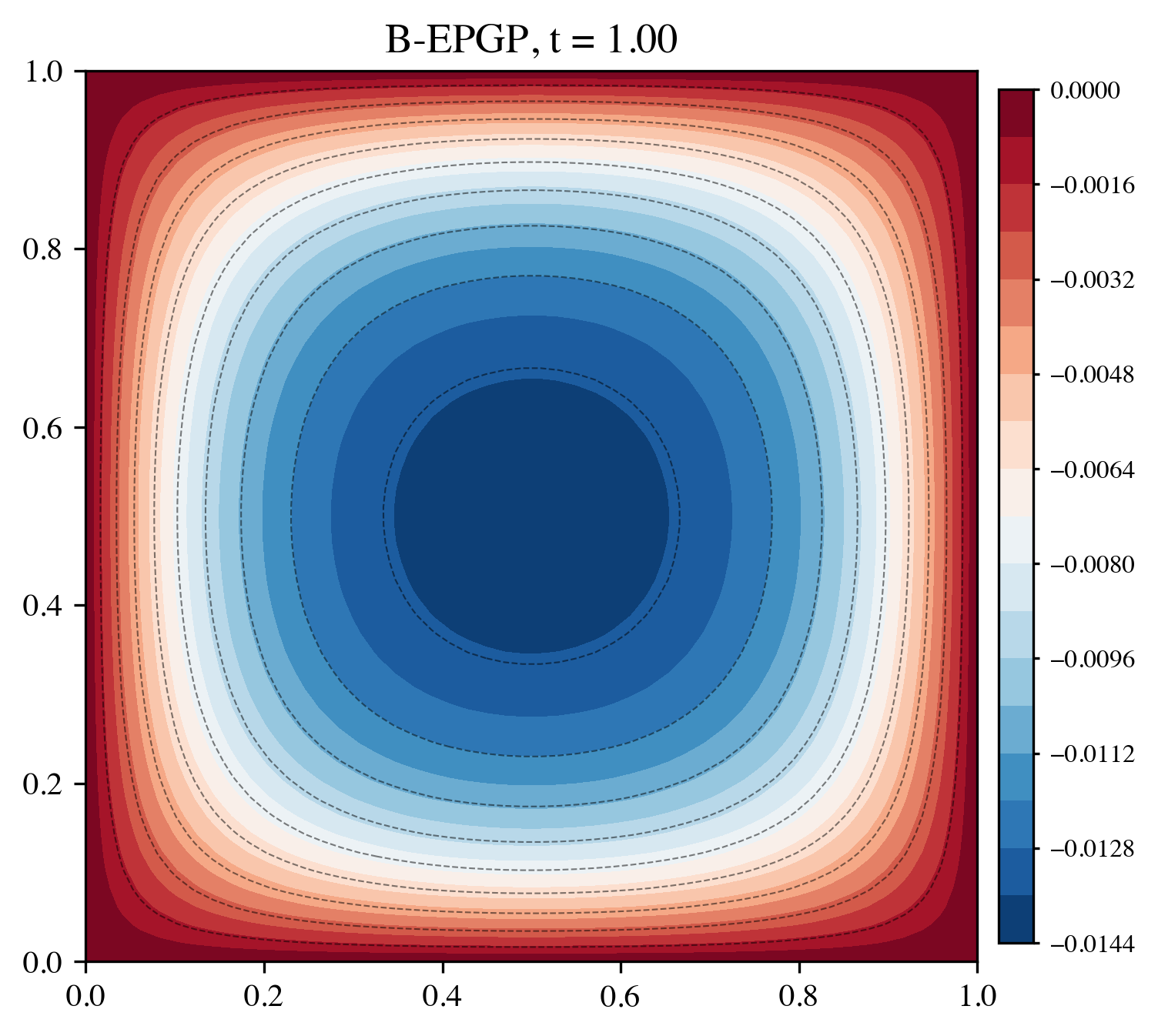}
    \caption{\tiny $t=1.0$,\\ B-EPGP}
  \end{subfigure}
   \\
  \begin{subfigure}[b]{0.15\textwidth}
    \centering
    \includegraphics[width=\textwidth]{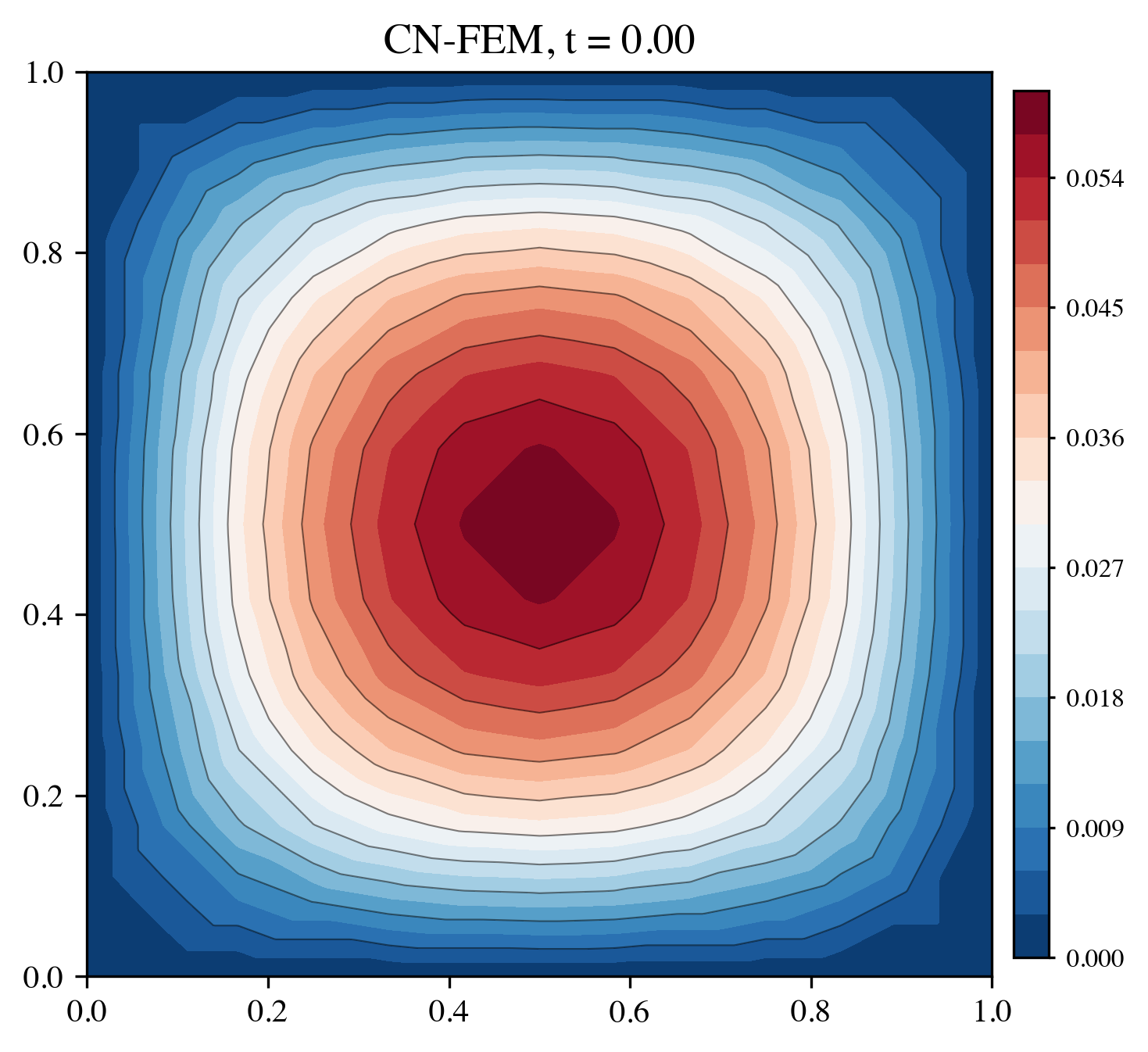}
    \caption{\tiny $t=0.0$,\\ CN-FEM}
  \end{subfigure}
  \hfill
  \begin{subfigure}[b]{0.15\textwidth}
    \centering
    \includegraphics[width=\textwidth]{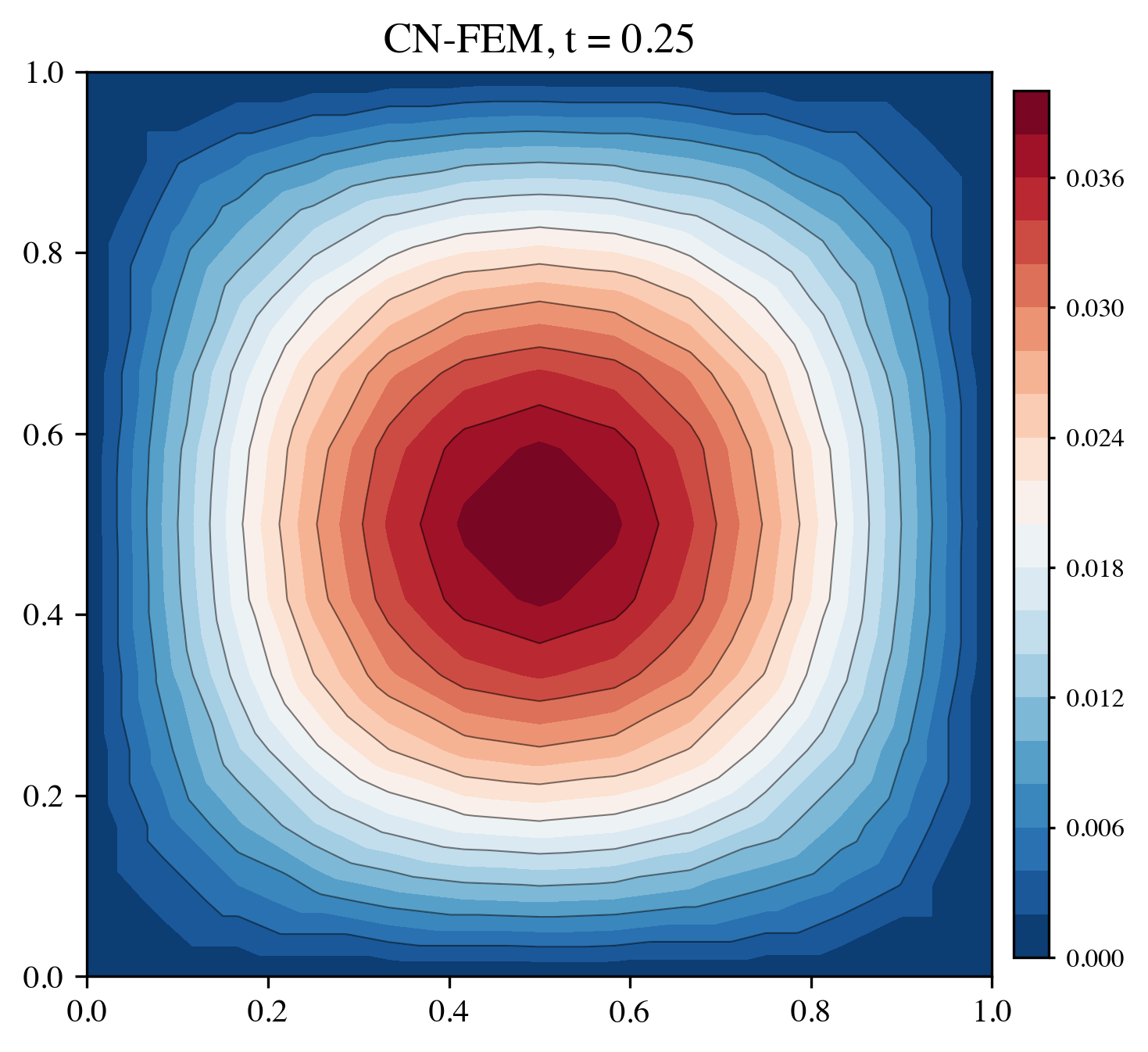}
    \caption{\tiny $t=0.25$,\\ CN-FEM}
  \end{subfigure}
  \hfill
  \begin{subfigure}[b]{0.15\textwidth}
    \centering
    \includegraphics[width=\textwidth]{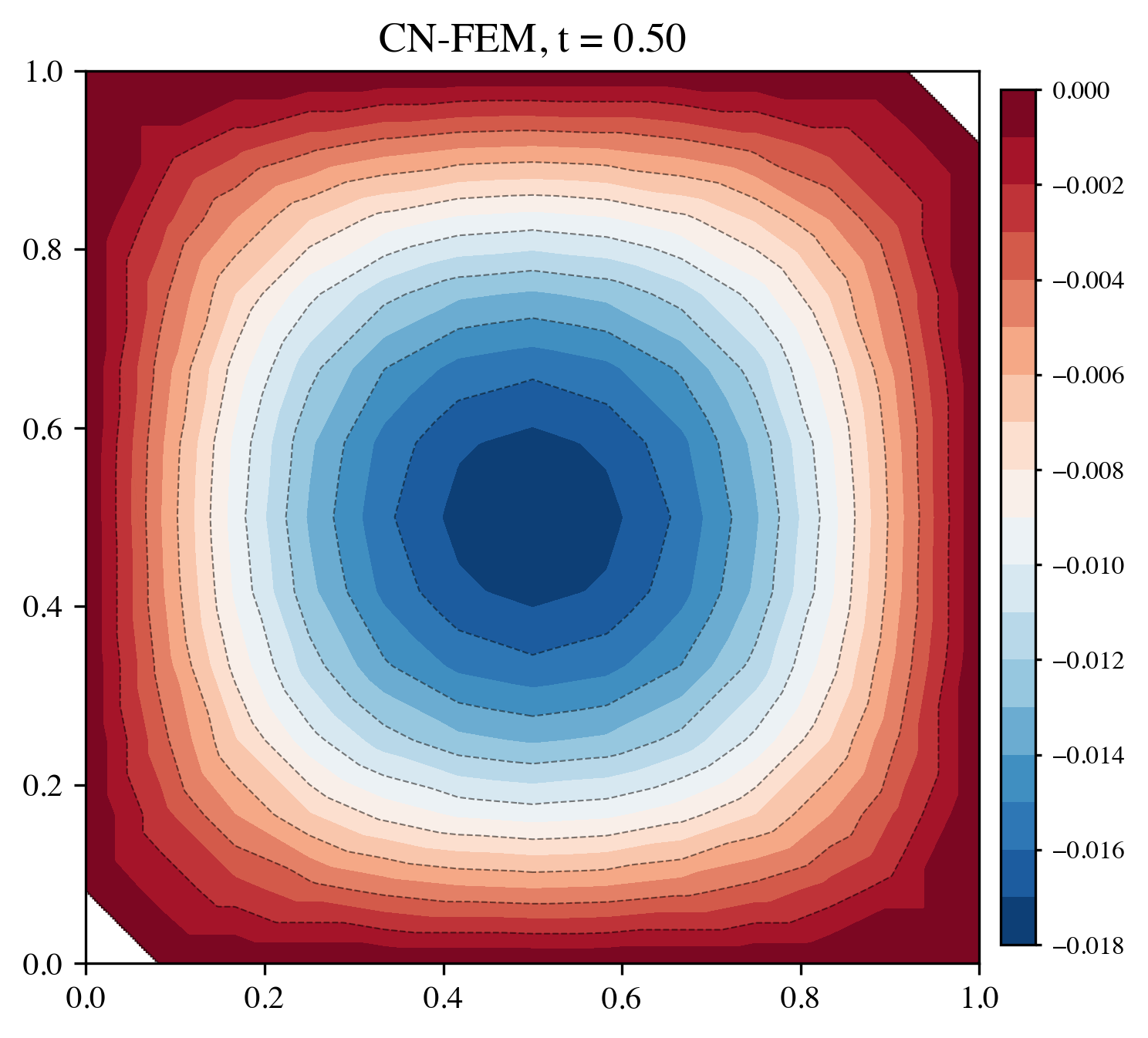}
    \caption{\tiny $t=0.50$,\\ CN-FEM}
  \end{subfigure}
  \hfill
  \begin{subfigure}[b]{0.15\textwidth}
    \centering
    \includegraphics[width=\textwidth]{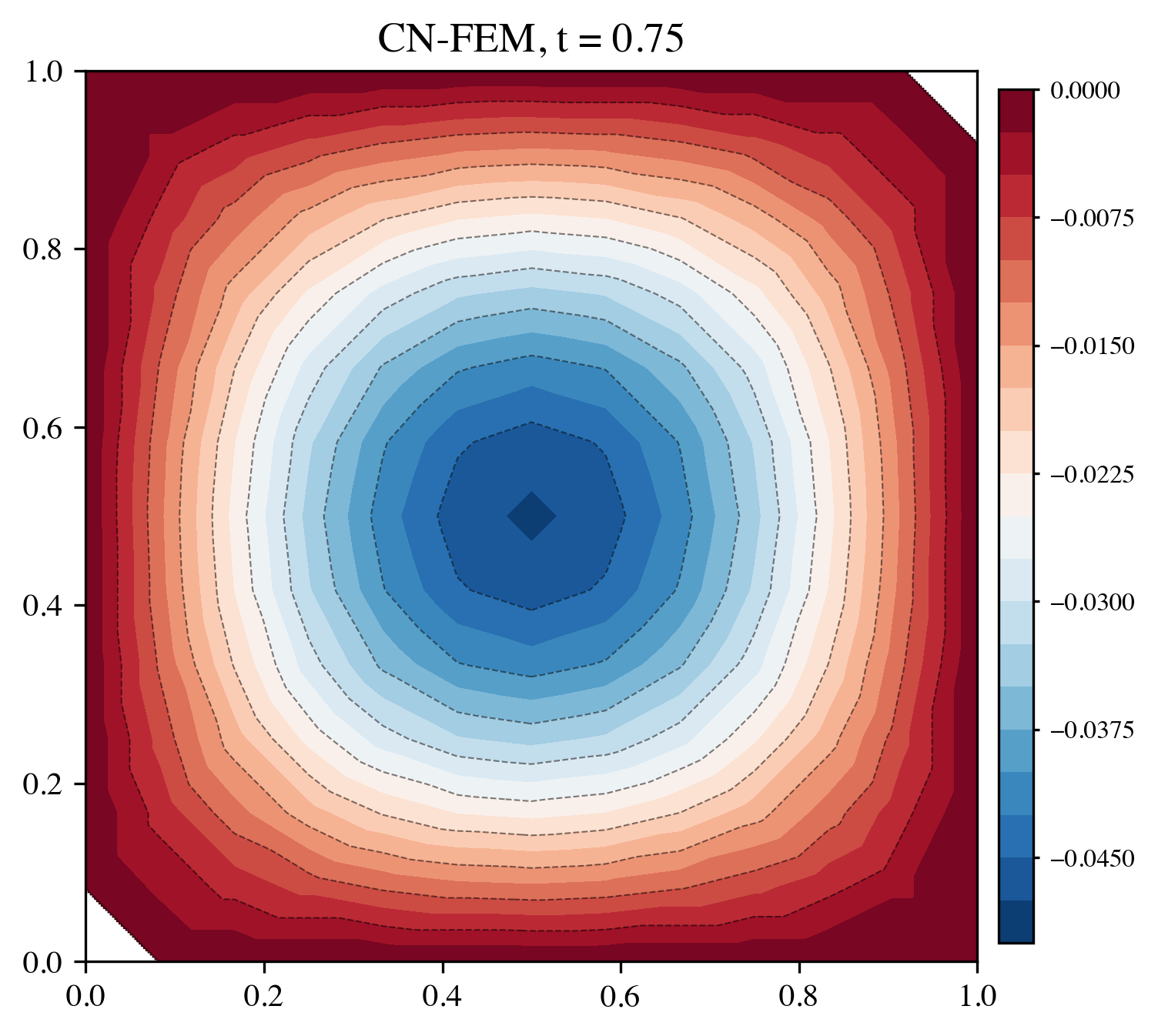}
    \caption{\tiny $t=0.75$,\\ CN-FEM}
  \end{subfigure}
  \hfill
  \begin{subfigure}[b]{0.15\textwidth}
    \centering
    \includegraphics[width=\textwidth]{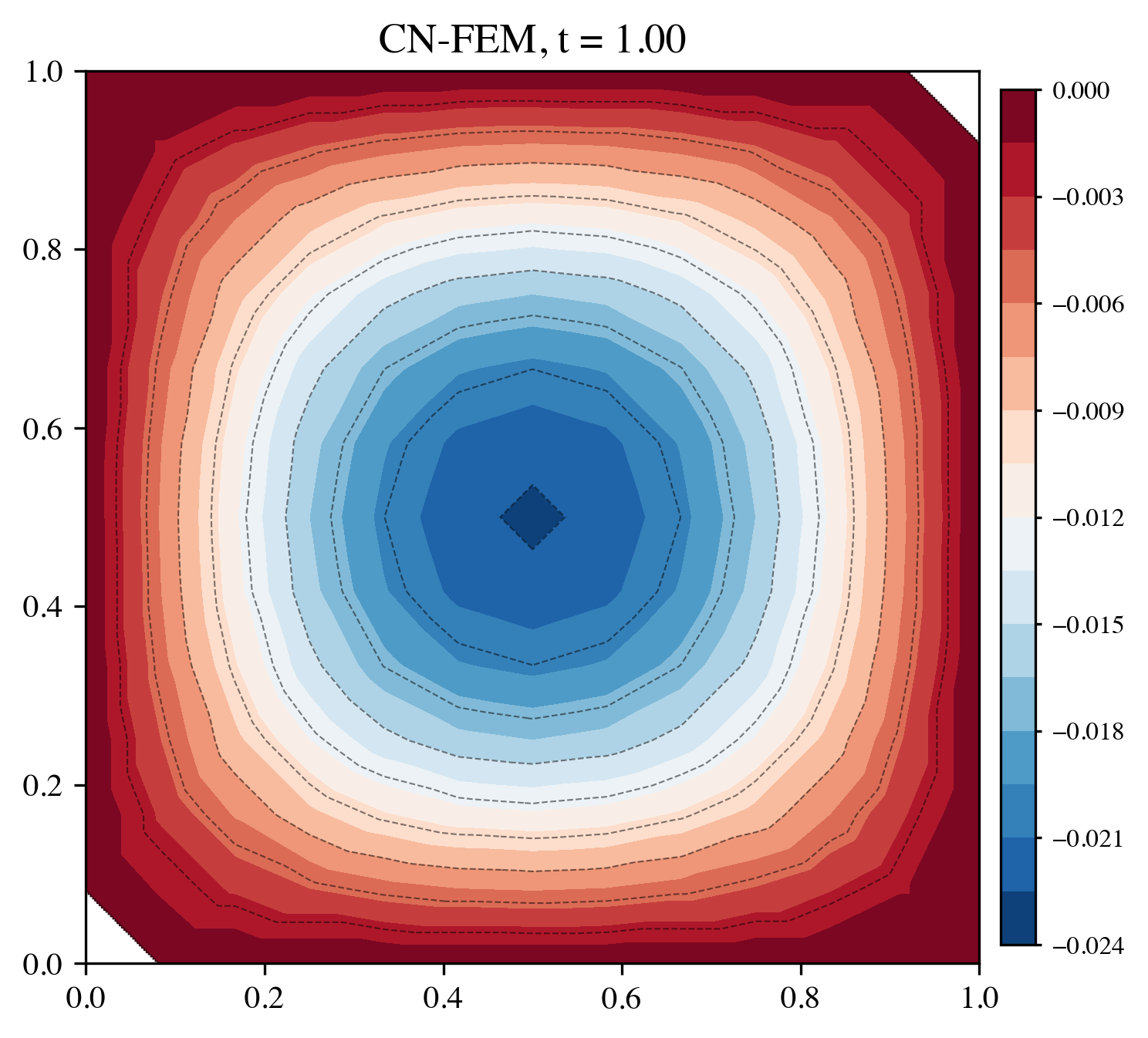}
    \caption{\tiny $t=1.0$,\\ CN-FEM}
  \end{subfigure}
  \caption{$2$-D contour plots of the solution with polynomial initial condition at  time stamps (\(t = 0.0, 0.25, 0.5, 0.75, 1.0 \)). Each row shows the reference solution (top), B-EPGP (middle), and CN-FEM approximation (bottom).}
  \label{figure_one}
  
\end{figure}

\begin{figure}[H]
  \vskip 0.2in
  \centering
  \begin{subfigure}[b]{0.15\textwidth}
    \centering
    \includegraphics[width=\textwidth]{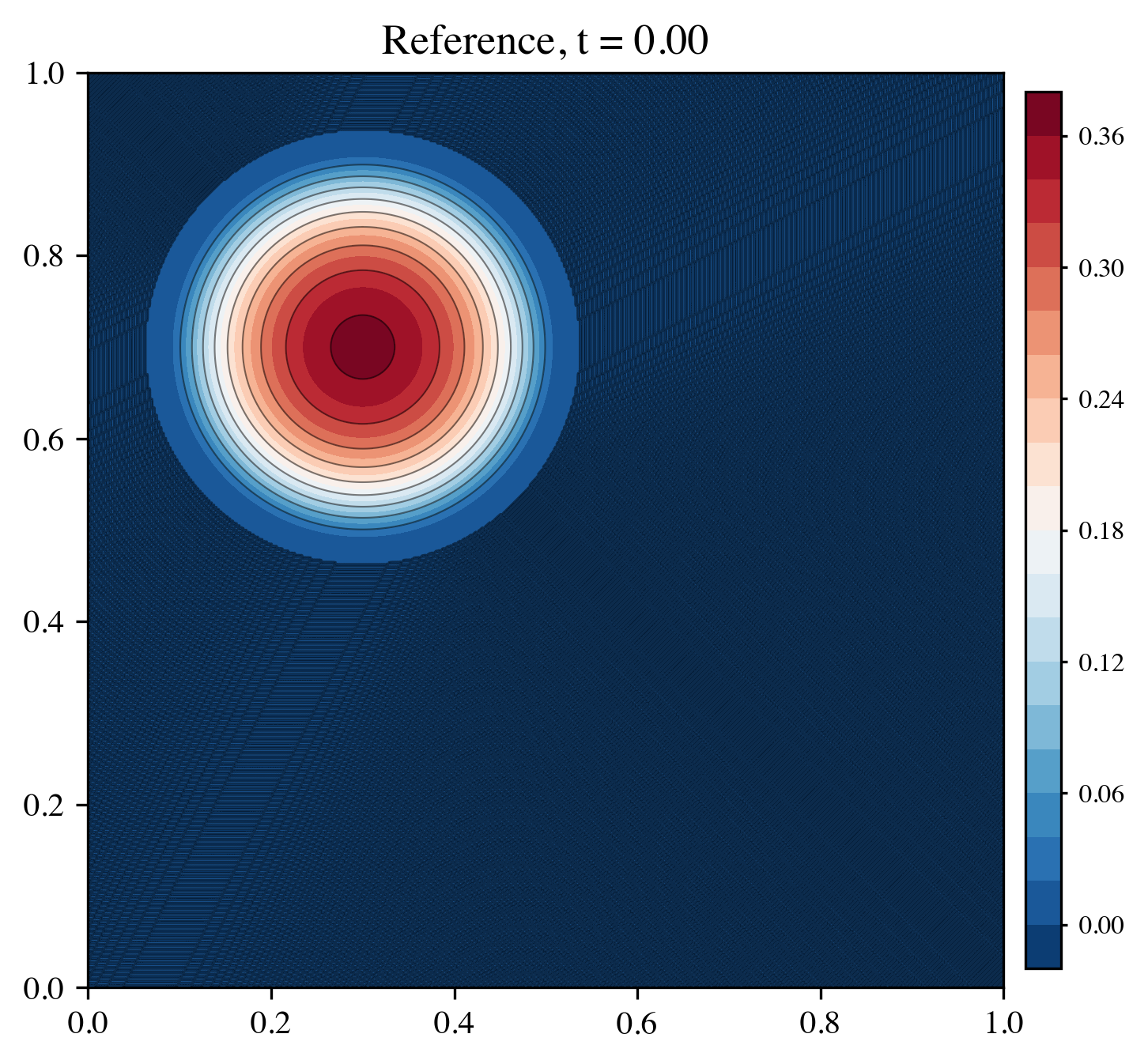}
    \caption{\tiny $t=0.0$,\\ Reference}
  \end{subfigure}
  \hfill
  \begin{subfigure}[b]{0.15\textwidth}
    \centering
    \includegraphics[width=\textwidth]{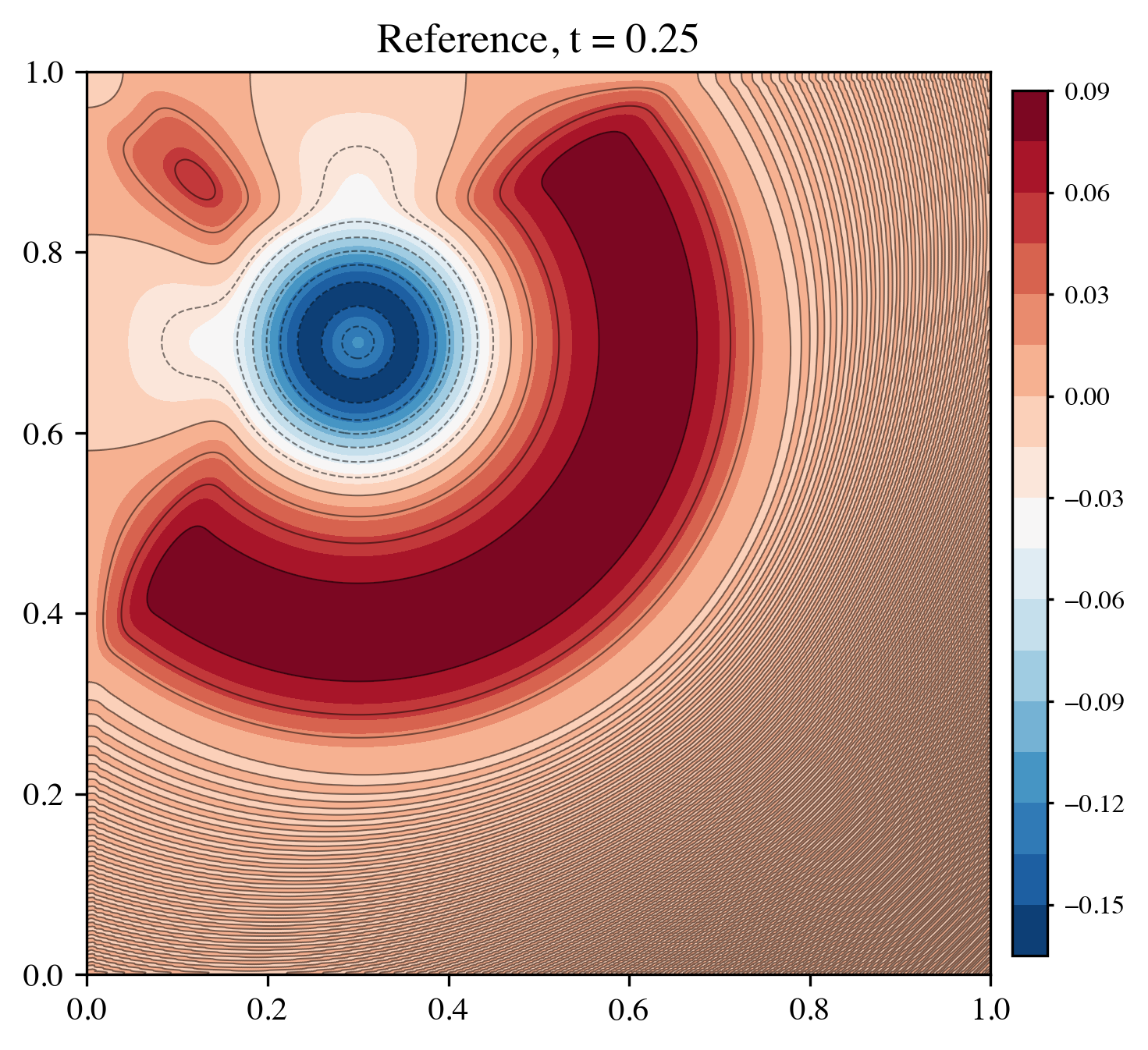}
    \caption{\tiny $t=0.25$,\\ Reference}
  \end{subfigure}
  \hfill
  \begin{subfigure}[b]{0.15\textwidth}
    \centering
    \includegraphics[width=\textwidth]{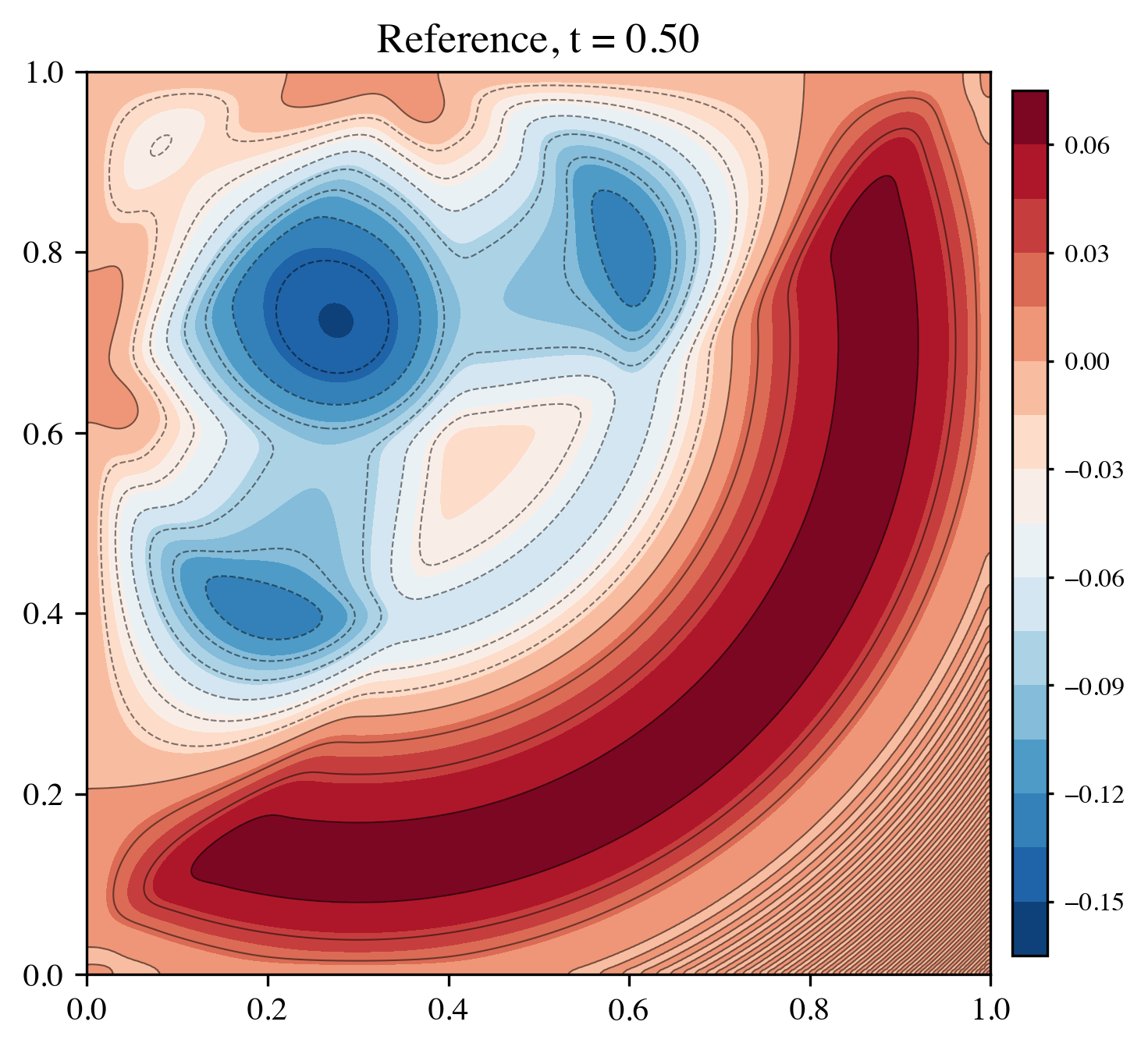}
    \caption{\tiny $t=0.50$,\\ Reference}
  \end{subfigure}
  \hfill
  \begin{subfigure}[b]{0.15\textwidth}
    \centering
    \includegraphics[width=\textwidth]{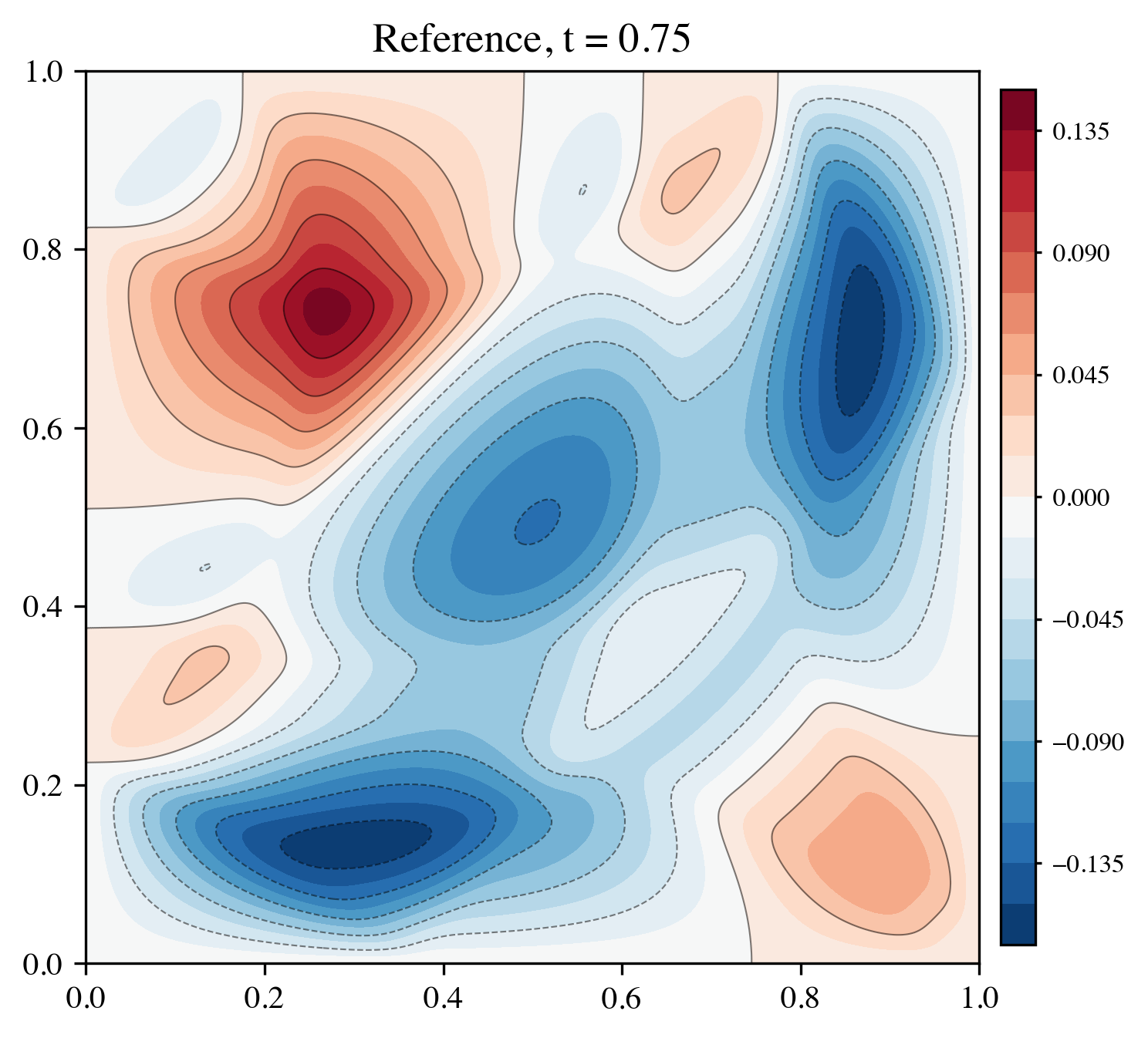}
    \caption{\tiny $t=0.75$,\\ Reference}
  \end{subfigure}
  \hfill
  \begin{subfigure}[b]{0.15\textwidth}
    \centering
    \includegraphics[width=\textwidth]{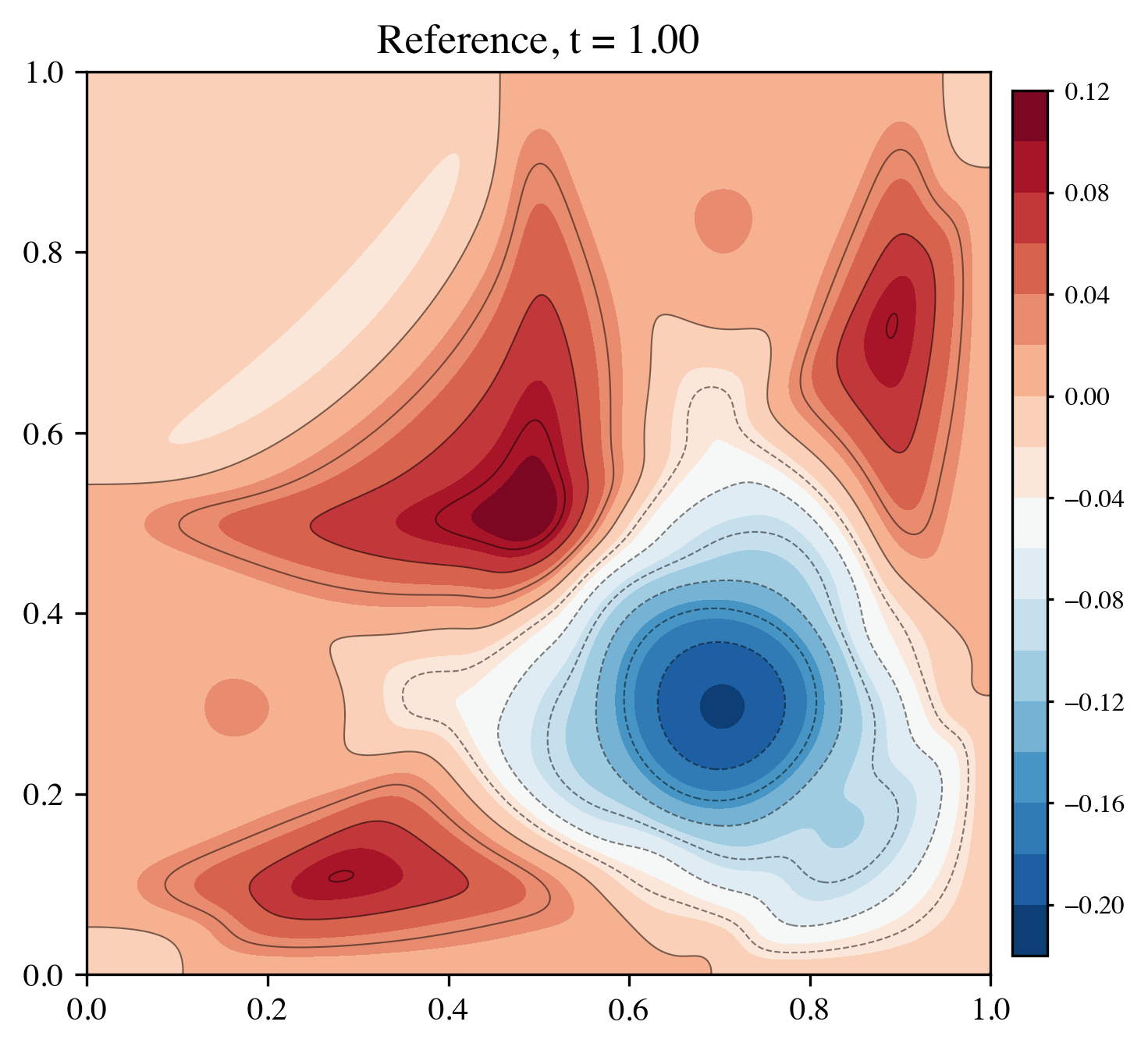}
    \caption{\tiny $t=1.0$,\\ Reference}
  \end{subfigure}
  \\
  \begin{subfigure}[b]{0.15\textwidth}
    \centering
    \includegraphics[width=\textwidth]{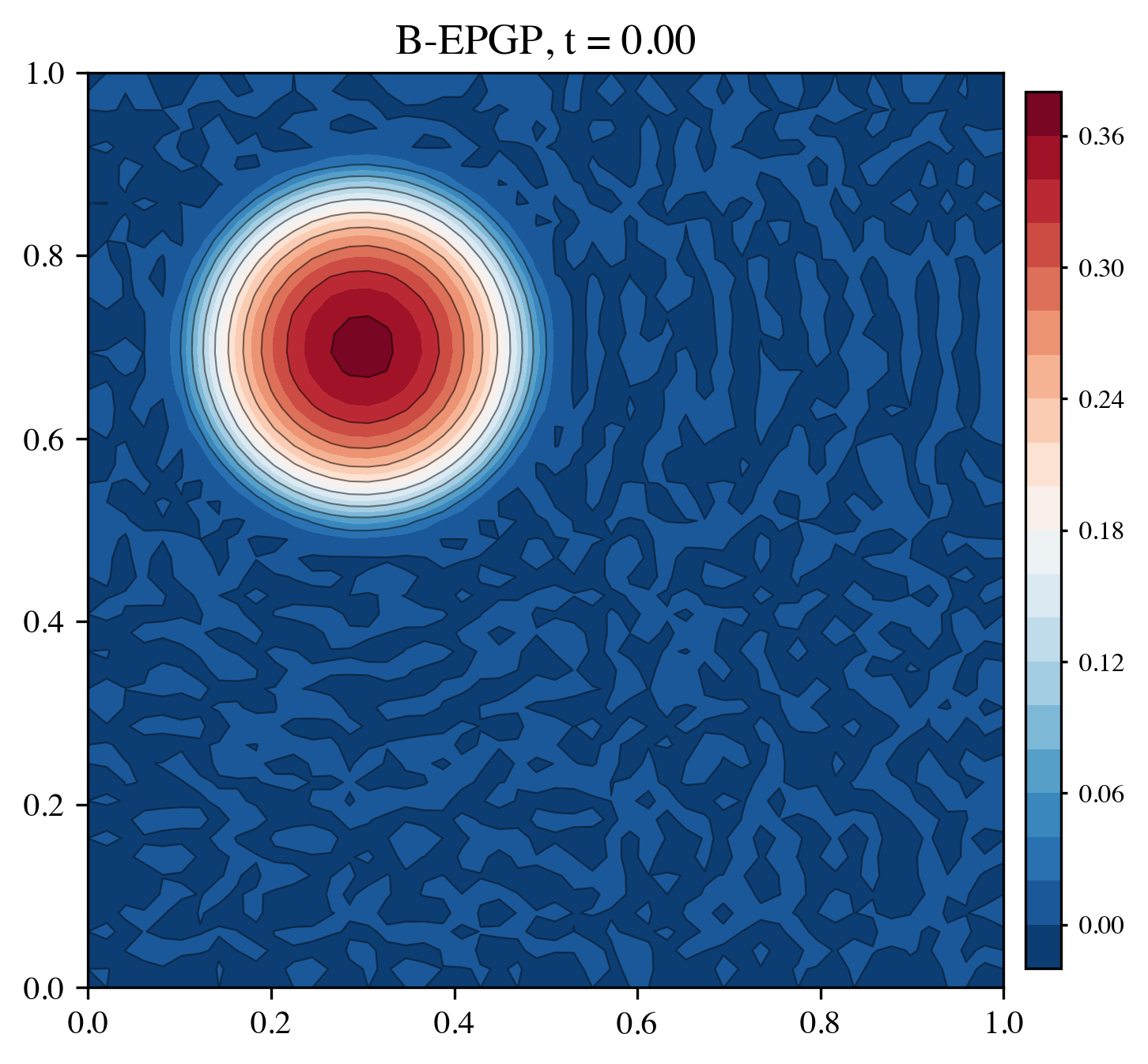}
    \caption{\tiny $t=0.0$,\\ B-EPGP}
  \end{subfigure}
  \hfill
  \begin{subfigure}[b]{0.15\textwidth}
    \centering
    \includegraphics[width=\textwidth]{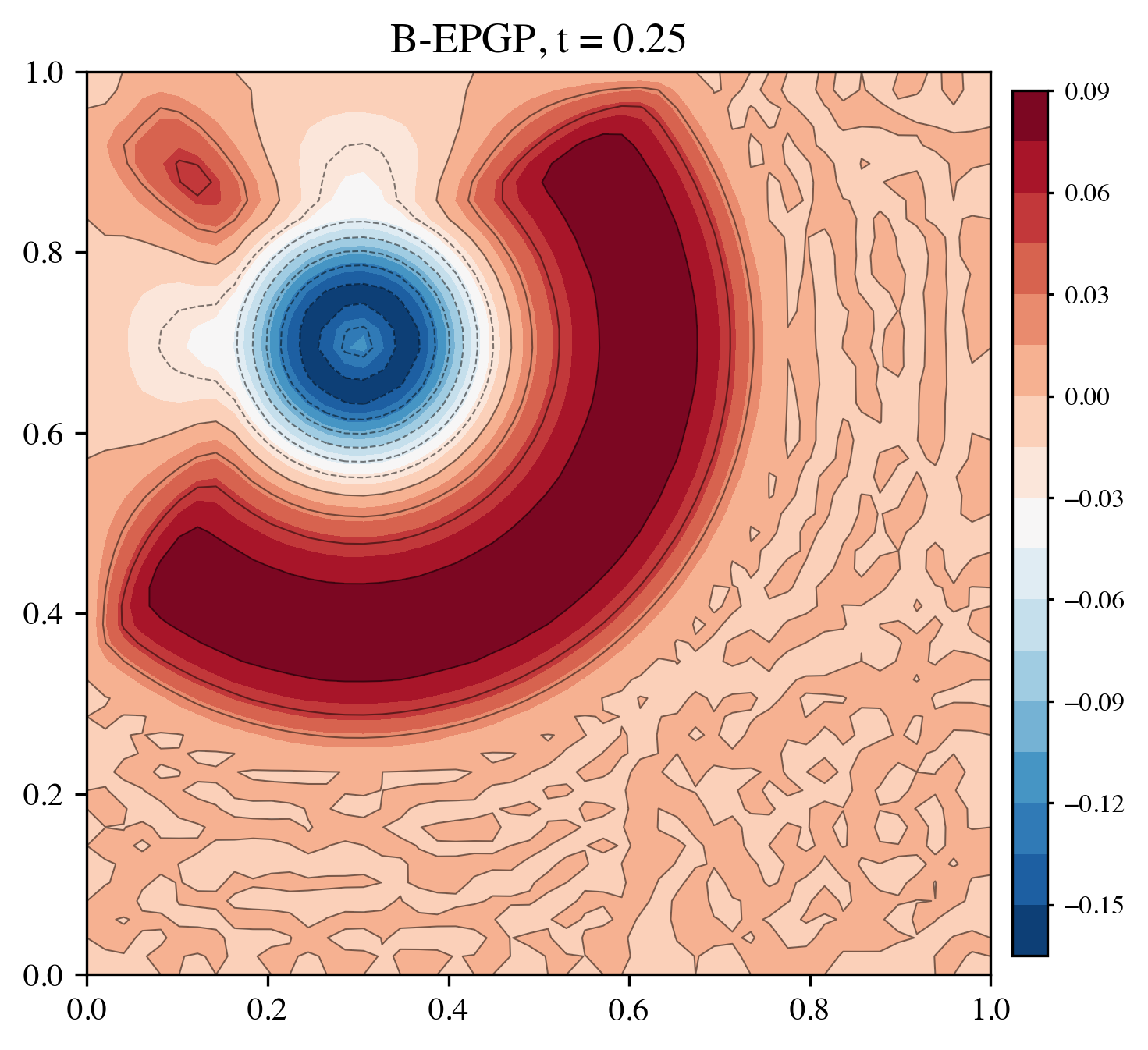}
    \caption{\tiny $t=0.25$,\\ B-EPGP}
  \end{subfigure}
  \hfill
  \begin{subfigure}[b]{0.15\textwidth}
    \centering
    \includegraphics[width=\textwidth]{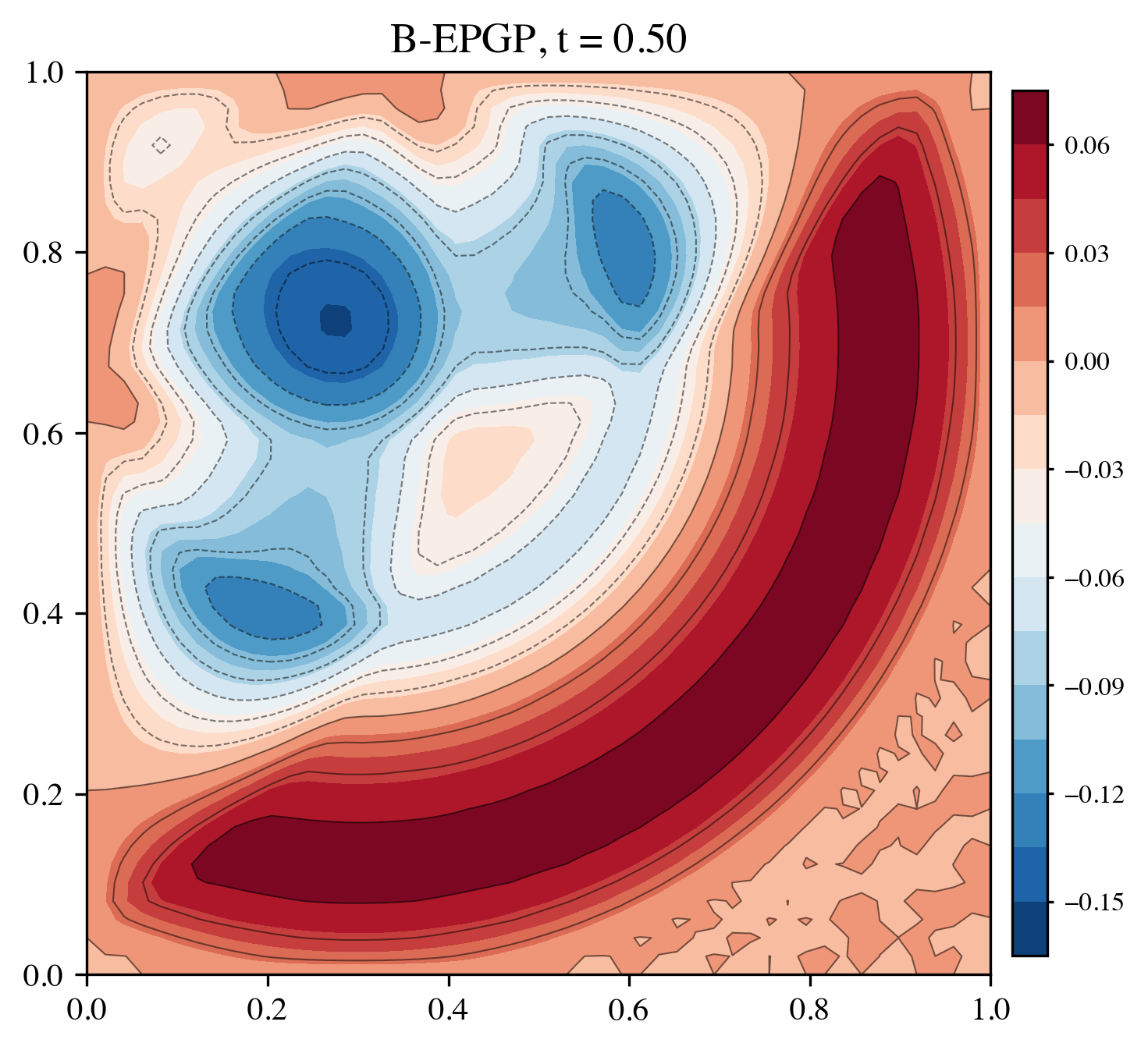}
    \caption{\tiny $t=0.50$,\\ B-EPGP}
  \end{subfigure}
  \hfill
  \begin{subfigure}[b]{0.15\textwidth}
    \centering
    \includegraphics[width=\textwidth]{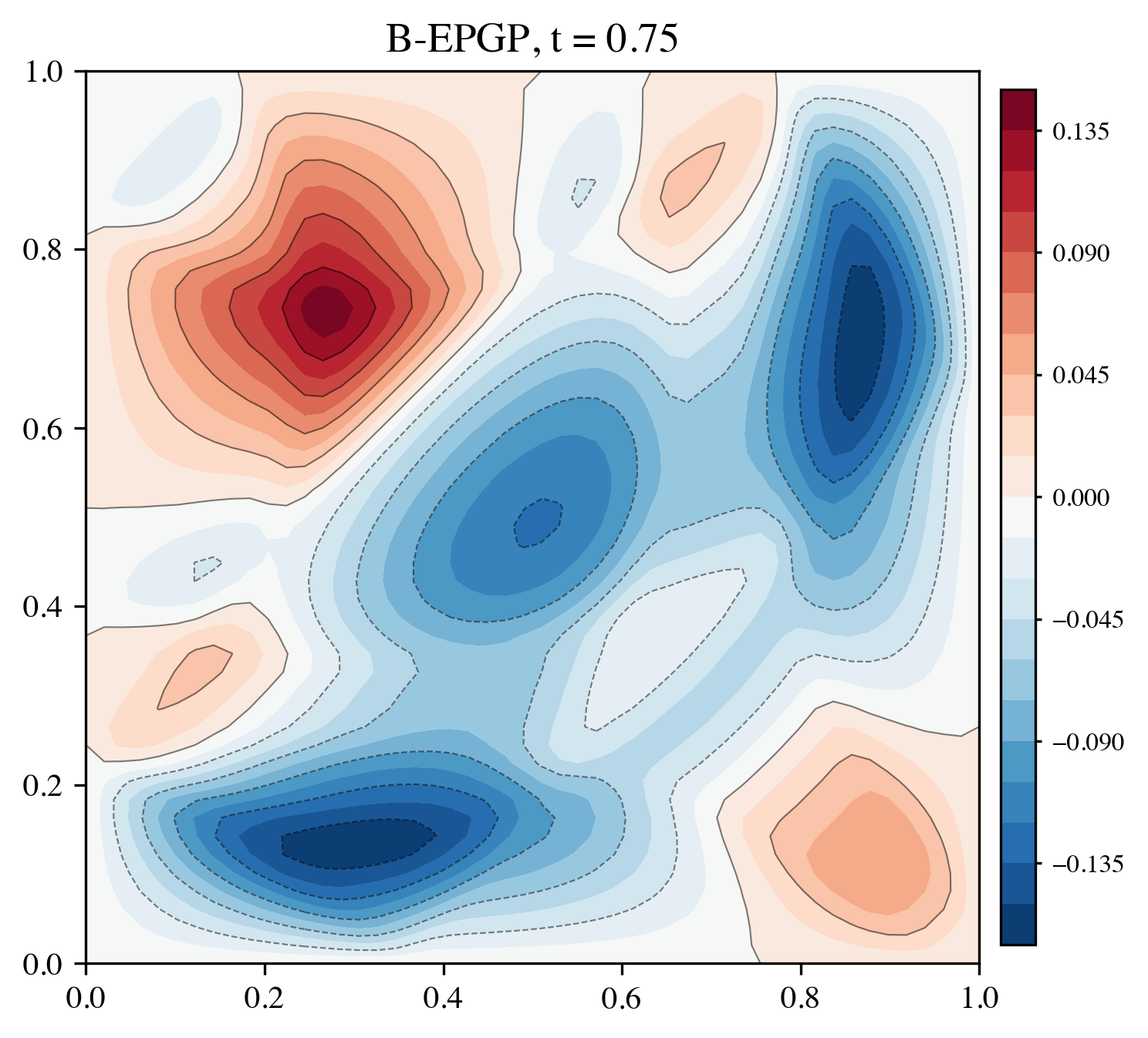}
    \caption{\tiny $t=0.75$,\\ B-EPGP}
  \end{subfigure}
  \hfill
  \begin{subfigure}[b]{0.15\textwidth}
    \centering
    \includegraphics[width=\textwidth]{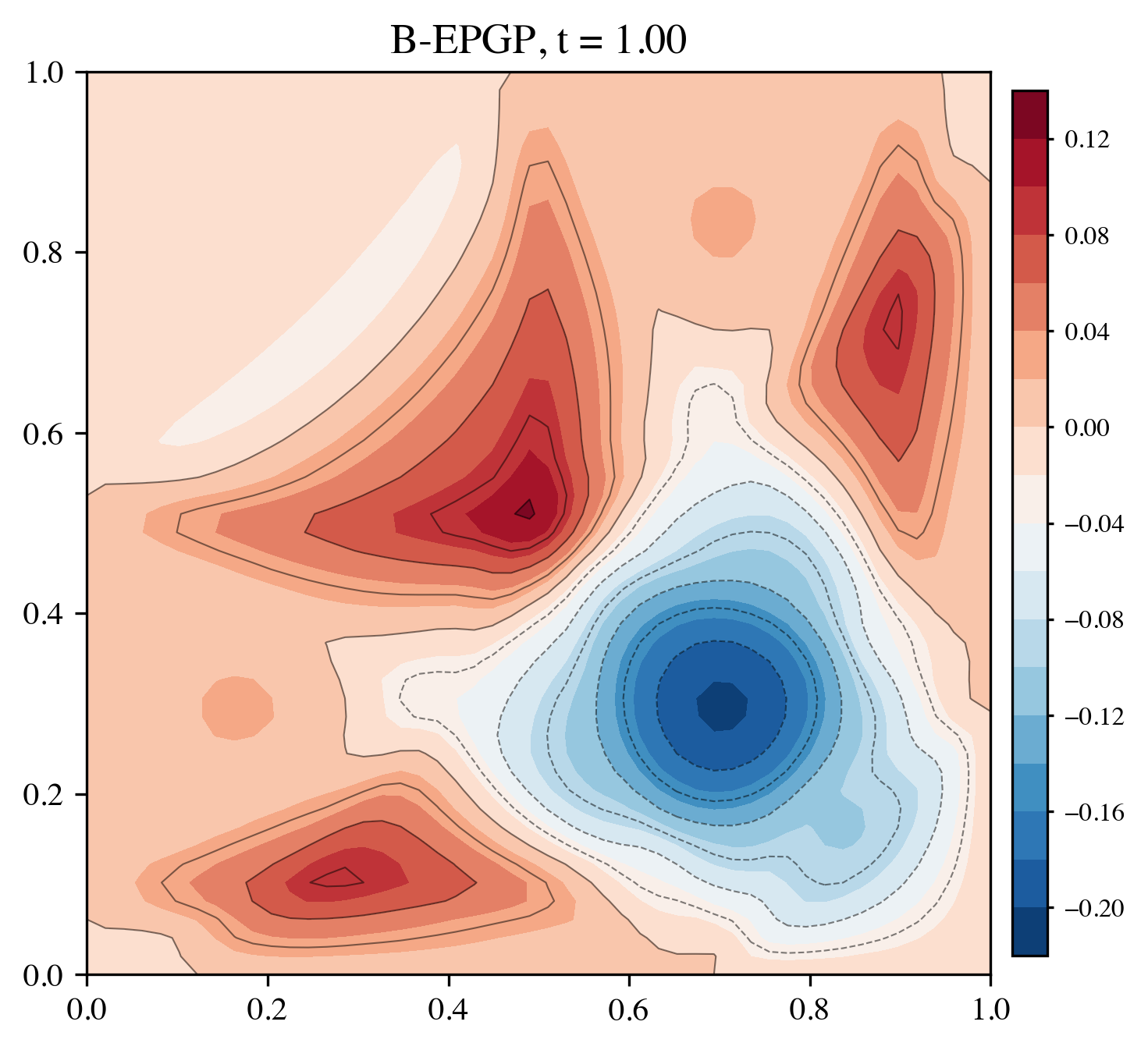}
    \caption{\tiny $t=1.0$,\\ B-EPGP}
  \end{subfigure}
   \\
  \begin{subfigure}[b]{0.15\textwidth}
    \centering
    \includegraphics[width=\textwidth]{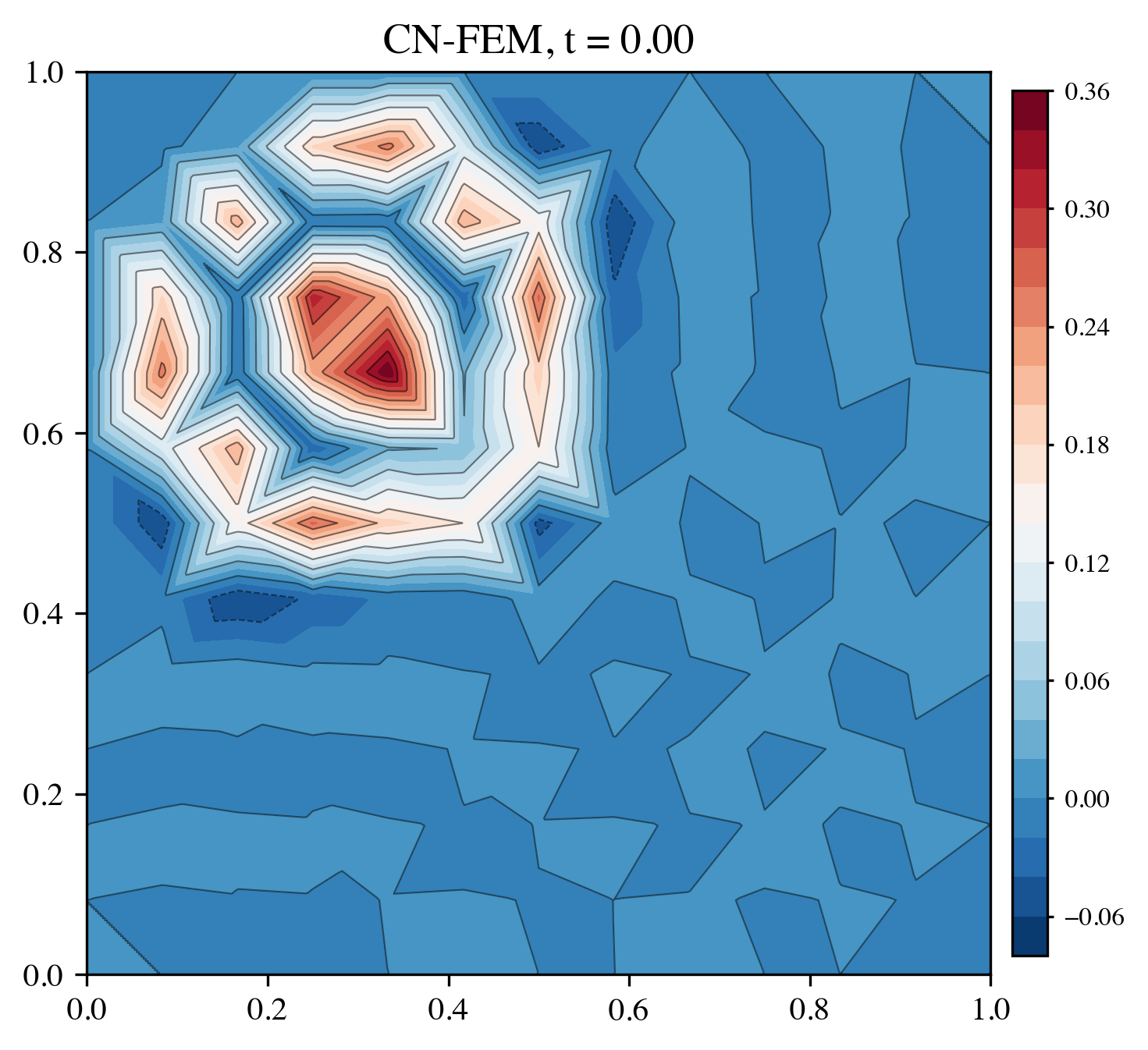}
    \caption{\tiny $t=0.0$,\\ CN-FEM}
  \end{subfigure}
  \hfill
  \begin{subfigure}[b]{0.15\textwidth}
    \centering
    \includegraphics[width=\textwidth]{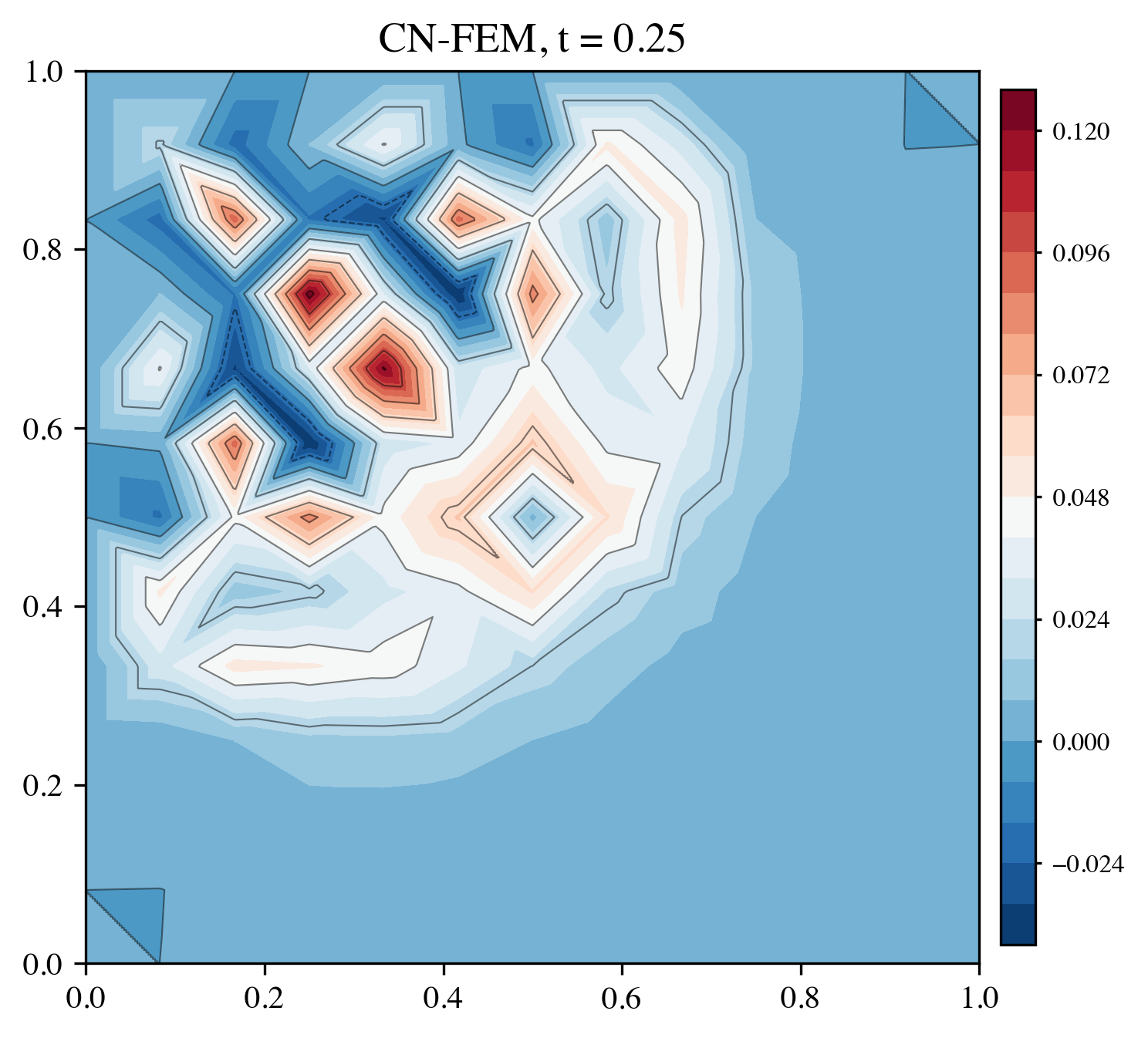}
    \caption{\tiny $t=0.25$,\\ CN-FEM}
  \end{subfigure}
  \hfill
  \begin{subfigure}[b]{0.15\textwidth}
    \centering
    \includegraphics[width=\textwidth]{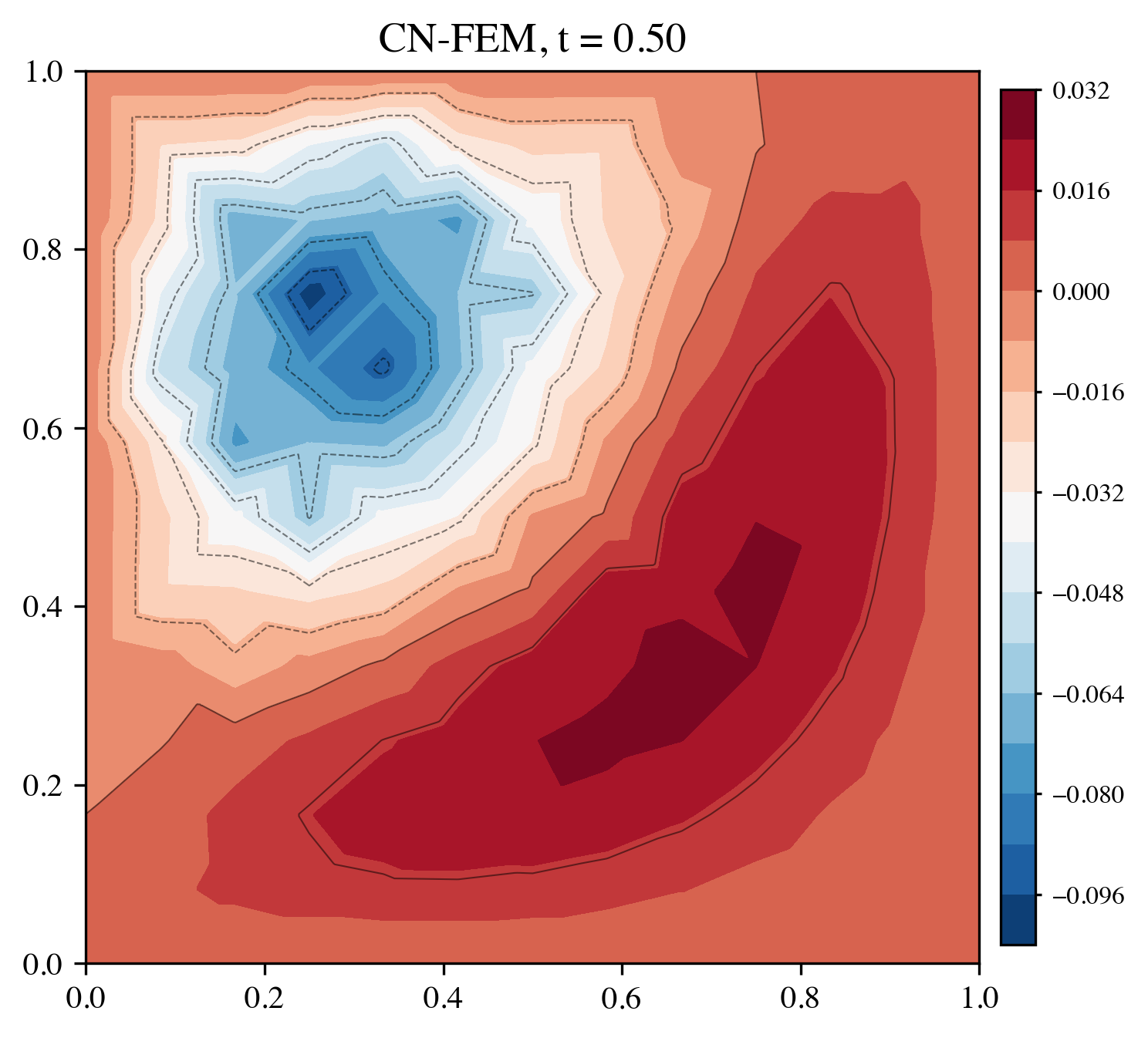}
    \caption{\tiny $t=0.50$,\\ CN-FEM}
  \end{subfigure}
  \hfill
  \begin{subfigure}[b]{0.15\textwidth}
    \centering
    \includegraphics[width=\textwidth]{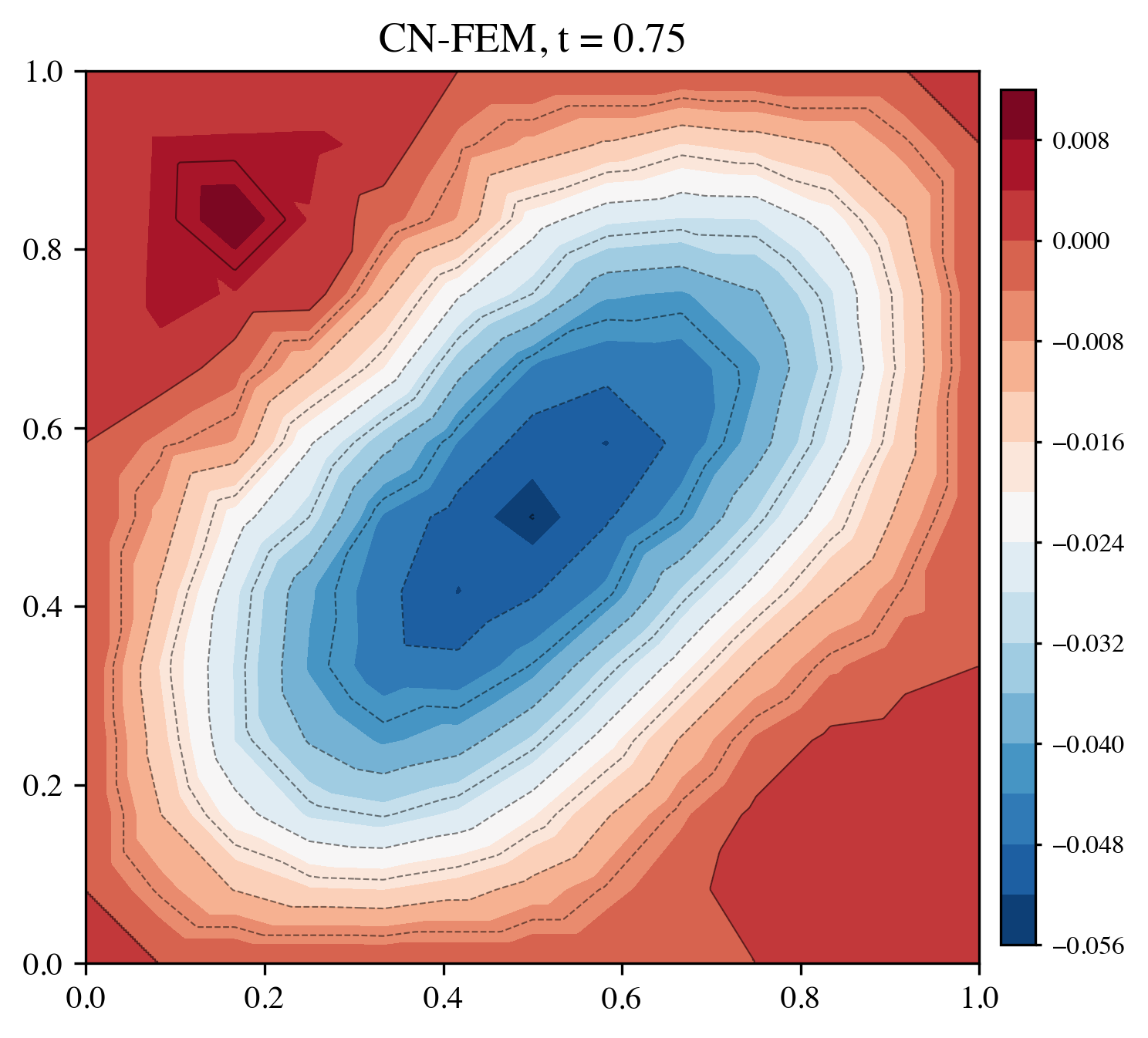}
    \caption{\tiny $t=0.75$,\\ CN-FEM}
  \end{subfigure}
  \hfill
  \begin{subfigure}[b]{0.15\textwidth}
    \centering
    \includegraphics[width=\textwidth]{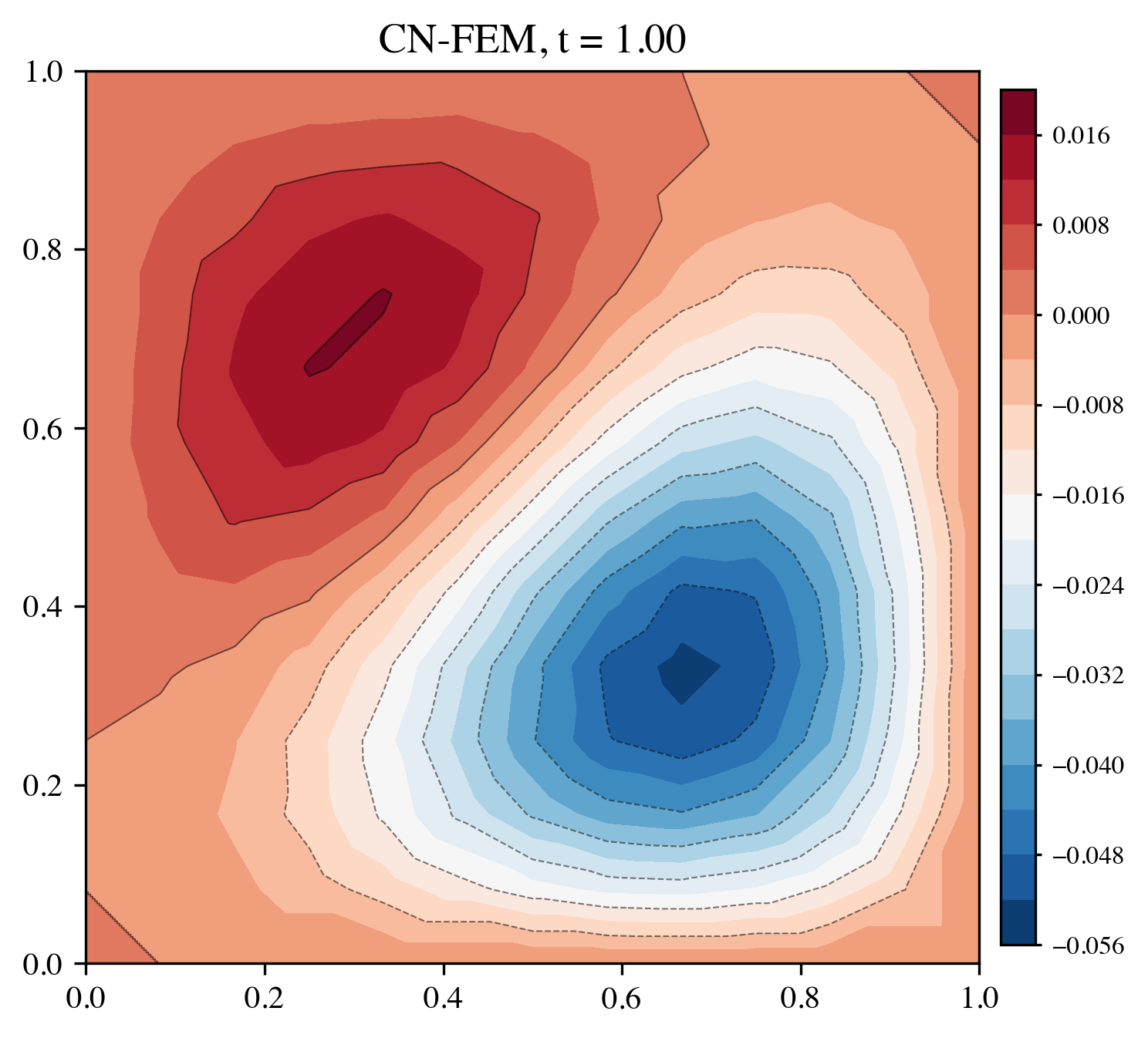}
    \caption{\tiny $t=1.0$,\\ CN-FEM}
  \end{subfigure}
  \caption{$2$-D contour plots of the solution with mollifier initial condition at time stamps (\(t = 0.0, 0.25, 0.5, 0.75, 1.0 \)). Each row shows the reference solution (top), B-EPGP (middle), and CN-FEM approximation (bottom).}
  \label{figure_two}
  
\end{figure}

\fi

\paragraph{\textbf{Mollifier}.}
The optimal $\lambda^{\ast} = 1.30\times10^{-3}$ yielded an effective DoF of approximately $1{,}600$, while the matched CN-FEM setup again used $1{,}573$ DoF. 
Under these settings, the space–time $L^{2}$-error decreased from $4.43\times10^{-2}$ for CN-FEM to $5.68\times10^{-4}$ for B-EPGP, corresponding to an $78.08\times$ improvement, with the relative error dropping from $75.31\%$ to $0.96\%$.
The maximum-in-time spatial $L^{2}$-error was likewise much smaller for B-EPGP, decreasing from $6.59\times10^{-2}$ for CN-FEM to $1.02\times10^{-3}$, with the relative error reduced from $79.93\%$ to $1.24\%$.
The GCV optimization converged in 1.332\,s, while the CN-FEM solve required 0.037\,s.

For visual comparison, all solutions were evaluated at the temporal grid points of the reference simulation to ensure temporal consistency. The reference solution was visualized on its native fine mesh of $400\times400$, while the CN FEM results were interpolated onto the same reference grid for direct pointwise comparison. In contrast, the B EPGP solution, owing to its smooth spectral representation, was evaluated at $50\times50$, which provided comparable visual accuracy while significantly reducing the evaluation cost. Our visual comparison and numerical error computation are done on consistent grids and time points, so any observed differences truly reflect method accuracy, not discretization differences.

Figures~\ref{figure_one} and \ref{figure_two} present snapshots of the spatiotemporal evolution of the wave field for the two initial conditions, polynomial and mollifier, at snapshot times $t = 0.0, 0.25, 0.5, 0.75, 1.0$. Each row corresponds respectively to the reference solution (top), the proposed B-EPGP approximation (middle), and the CN-FEM solution (bottom).

In the polynomial case, the wave fields exhibits a smooth, symmetric standing-wave pattern centered within the domain. The B-EPGP reconstruction reproduces the reference amplitude and phase with remarkable fidelity across all time frames. In contrast, the CN-FEM solution preserves the overall modal structure but displays gradual amplitude damping and a slight phase delay beyond $t=0.5$, visible as small shift of the contour transitions near the extrema of displacement. The smoother and less contrasted appearance of the CN-FEM field reflects the mild numerical dissipation inherent in the implicit Crank–Nicolson time integration. These visual patterns are consistent with the quantitative results in Table \ref{tab:l2_space_time}, where the relative space-time error decreases from $32\%$ for CN-FEM to $0.1\%$ for B-EPGP, confirming an improvement of about three orders of magnitude in accuracy. 

In the mollifier case, the initial displacement forms a sharply localized bump centered at $(0.3, 0.7)$, generating an outward moving circular disturbance that interacts with the domain boundaries over time. The reference solution shows smooth, symmetric wavefronts that expand uniformly and reflects smoothly at the boundaries. The B-EPGP approximation captures these dynamics with notable precision, maintaining amplitude, phase, and propagation speed in close agreement with the reference throughout the experiment. The reflected patterns stay well organized, and the overall energy is preserved. Small oscillations appear behind the main wavefront due to the finite trigonometric basis, but they remain localized and have negligible effect on the dominant propagation. By contrast, the CN FEM solution captures the general wave behavior but exhibits visible numerical diffusion, particularly after multiple reflections. The wavefront becomes broader and less defined, and the amplitude decays more rapidly after each reflection. By $t=1.0$, the CN-FEM contours appear noticeably blurred and slightly phase-shifted relative to the reference, whereas B-EPGP retains a crisp and symmetric front. These visual features align with the quantitative results in Table \ref{tab:l2_space_time}, where the relative space-time $L^2$-error decreases from about $75\%$ for the CN-FEM to $0.96\%$ for B-EPGP, indicating an improvement of nearly two orders of magnitude in accuracy. 

A similar trend can be seen in A consistent pattern is also observed in Table~\ref{tab:linf_time}, where the maximum in time spatial error shows comparable reductions for both initial conditions. This agreement between the $L^{^2}(0,T;L^{2}(\Omega))$-error and $L^{\infty}(0,T;L^{2}(\Omega))$-error measures confirms the robustness and stability of the B-EPGP surrogate across different norms.

Across both initial conditions, B-EPGP consistently preserves phase alignment and amplitude with minimal dissipation while maintaining computational efficiency. The method reproduces smooth standing-wave structures for the polynomial case and localized propagating features for the mollifier case with consistently high accuracy. The CN-FEM solutions, though stable and physically plausible, exhibit mild numerical diffusion and phase delay typical of implicit integration. These visual trends reinforce the quantitative results, confirming that B-EPGP provides a more accurate and stable representation of the true wave dynamics throughout the simulation period, with noticeably sharper and more energy-preserving contours across all time frames.

\section{Conclusion}\label{sec:conclusion}

The comparative analysis between B-EPGP and the Crank Nicolson FEM demonstrates the effectiveness of operator informed Gaussian process models as efficient and accurate surrogates for partial differential equations. 
Under matched degrees of freedom, B-EPGP consistently attains smaller errors while maintaining comparable or shorter computation times. 
For the smooth polynomial case, the errors improve by approximately three orders of magnitude (about $316\times$), and for the localized mollifier case, the reductions reach nearly two orders of magnitude (about $78\times$), confirming consistent gains across different measures. 
These results demonstrate that the proposed method can simulate complex wave dynamics with higher precision and reduced numerical dissipation compared to classical finite element discretizations. The advantage of B-EPGP lies in its construction, where each basis function satisfies the governing wave equation and boundary conditions exactly, embedding the operator physics directly into the representation space. This property enables the surrogate to achieve deterministic accuracy with probabilistic flexibility, bridging the gap between numerical precision and data-driven inference. Overall, these findings confirm that B-EPGP provides a viable and interpretable alternative to traditional solvers for linear wave problems. By combining analytical structure with statistical regularization, it achieves high accuracy, computational efficiency, and inherent uncertainty quantification.

\bibliographystyle{abbrv}
\bibliography{refs}

\end{document}